\documentclass[11pt]{article}

\usepackage[usenames,dvipsnames]{xcolor}
\usepackage{syntaxfest2021}

\usepackage{amssymb}
\usepackage{amsmath}

\usepackage{times}
\usepackage{url}

\usepackage[normalem]{ulem}

\usepackage{multirow}
\usepackage{booktabs}

\usepackage{latexsym}
\usepackage{tikz}

\newcommand \mymr[3] { \multirow{#1}{*}{\parbox{#2}{#3}} }

\usepackage{todonotes}

\usepackage[frozencache=true,cachedir=minted-cache]{minted}
\usepackage{tcolorbox}
\usepackage{etoolbox}
\BeforeBeginEnvironment{minted}{\begin{tcolorbox}}
\AfterEndEnvironment{minted}{\end{tcolorbox}}

\usepackage{newfloat}
\DeclareFloatingEnvironment[
    fileext=los,
    listname={List of Codes},
    name=Code,
    placement=tbhp,
]{code}

\newcommand \module[1]  {{\em #1}}

\newcommand \linktogithub[1] {\url{https://github.com/LAL-project/#1.git}}

\usetikzlibrary{arrows.meta,patterns}
\usetikzlibrary{ipe} 

\tikzstyle{ipe stylesheet} = [
  ipe import,
  even odd rule,
  line join=round,
  line cap=butt,
  ipe pen normal/.style={line width=0.4},
  ipe pen heavier/.style={line width=0.8},
  ipe pen fat/.style={line width=1.2},
  ipe pen ultrafat/.style={line width=2},
  ipe pen normal,
  ipe mark normal/.style={ipe mark scale=3},
  ipe mark large/.style={ipe mark scale=5},
  ipe mark small/.style={ipe mark scale=2},
  ipe mark tiny/.style={ipe mark scale=1.1},
  ipe mark normal,
  /pgf/arrow keys/.cd,
  ipe arrow normal/.style={scale=7},
  ipe arrow large/.style={scale=10},
  ipe arrow small/.style={scale=5},
  ipe arrow tiny/.style={scale=3},
  ipe arrow normal,
  /tikz/.cd,
  ipe arrows, 
  <->/.tip = ipe normal,
  ipe dash normal/.style={dash pattern=},
  ipe dash dotted/.style={dash pattern=on 1bp off 3bp},
  ipe dash dashed/.style={dash pattern=on 4bp off 4bp},
  ipe dash dash dotted/.style={dash pattern=on 4bp off 2bp on 1bp off 2bp},
  ipe dash dash dot dotted/.style={dash pattern=on 4bp off 2bp on 1bp off 2bp on 1bp off 2bp},
  ipe dash normal,
  ipe node/.append style={font=\normalsize},
  ipe stretch normal/.style={ipe node stretch=1},
  ipe stretch normal,
  ipe opacity 10/.style={opacity=0.1},
  ipe opacity 30/.style={opacity=0.3},
  ipe opacity 50/.style={opacity=0.5},
  ipe opacity 75/.style={opacity=0.75},
  ipe opacity opaque/.style={opacity=1},
  ipe opacity opaque,
]

\definecolor{red}{rgb}{1,0,0}
\definecolor{blue}{rgb}{0,0,1}
\definecolor{green}{rgb}{0,1,0}
\definecolor{yellow}{rgb}{1,1,0}
\definecolor{orange}{rgb}{1,0.647,0}
\definecolor{gold}{rgb}{1,0.843,0}
\definecolor{purple}{rgb}{0.627,0.125,0.941}
\definecolor{gray}{rgb}{0.745,0.745,0.745}
\definecolor{brown}{rgb}{0.647,0.165,0.165}
\definecolor{navy}{rgb}{0,0,0.502}
\definecolor{pink}{rgb}{1,0.753,0.796}
\definecolor{seagreen}{rgb}{0.18,0.545,0.341}
\definecolor{turquoise}{rgb}{0.251,0.878,0.816}
\definecolor{violet}{rgb}{0.933,0.51,0.933}
\definecolor{darkblue}{rgb}{0,0,0.545}
\definecolor{darkcyan}{rgb}{0,0.545,0.545}
\definecolor{darkgray}{rgb}{0.663,0.663,0.663}
\definecolor{darkgreen}{rgb}{0,0.392,0}
\definecolor{darkmagenta}{rgb}{0.545,0,0.545}
\definecolor{darkorange}{rgb}{1,0.549,0}
\definecolor{darkred}{rgb}{0.545,0,0}
\definecolor{lightblue}{rgb}{0.678,0.847,0.902}
\definecolor{lightcyan}{rgb}{0.878,1,1}
\definecolor{lightgray}{rgb}{0.827,0.827,0.827}
\definecolor{lightgreen}{rgb}{0.565,0.933,0.565}
\definecolor{lightyellow}{rgb}{1,1,0.878}
\definecolor{black}{rgb}{0,0,0}
\definecolor{white}{rgb}{1,1,1}
\syntaxfestfinalcopy

\title{The Linear Arrangement Library. A new tool for research on syntactic dependency structures.}

\author{
	Llu\'is Alemany-Puig \\
	{\tt\small lluis.alemany.puig@upc.edu} \\
	\\
	\And
	Juan Luis Esteban \\
	{\tt\small esteban@cs.upc.edu} \\
	{Universitat Polit\`ecnica de Catalunya} \\
	{Jordi Girona 1-3}\\
	{08034 Barcelona, Catalonia, Spain}
	\And
	Ramon Ferrer-i-Cancho \\
	{\tt\small ramon.ferrer@upc.edu} \\
}

\begin{document}

\maketitle

\begin{abstract}
The new and growing field of Quantitative Dependency Syntax has emerged at the crossroads between Dependency Syntax and Quantitative Linguistics. One of the main concerns in this field is the statistical patterns of syntactic dependency structures. These structures, grouped in treebanks, are the source for statistical analyses in these and related areas; dozens of scores devised over the years are the tools of a new industry to search for patterns and perform other sorts of analyses. The plethora of such metrics and their increasing complexity require sharing the source code of the programs used to perform such analyses. However, such code is not often shared with the scientific community or is tested following unknown standards. Here we present a new open-source tool, the Linear Arrangement Library (LAL), which caters to the needs of, especially, inexperienced programmers. This tool enables the calculation of these metrics on single syntactic dependency structures, treebanks, and collection of treebanks, grounded on ease of use and yet with great flexibility. LAL has been designed to be efficient, easy to use (while satisfying the needs of all levels of programming expertise), reliable (thanks to thorough testing), and to unite research from different traditions, geographic areas, and research fields.
\end{abstract}

\section{Introduction}

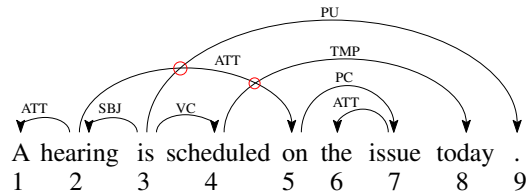
\begin{figure}
	\centering
\scalebox{0.9}{
\begin{tikzpicture}[ipe stylesheet]
  \node[ipe node]
     at (24, 140) {A};
  \node[ipe node]
     at (24, 128) {1};
  \node[ipe node]
     at (36, 140) { hearing};
  \node[ipe node]
     at (48, 128) {2};
  \node[ipe node]
     at (76, 140) { is};
  \node[ipe node]
     at (76, 128) {3};
  \node[ipe node]
     at (88, 140) { scheduled};
  \node[ipe node]
     at (104, 128) {4};
  \node[ipe node]
     at (136, 140) { on};
  \node[ipe node]
     at (136, 128) {5};
  \node[ipe node]
     at (152, 140) { the};
  \node[ipe node]
     at (156, 128) {6};
  \node[ipe node]
     at (172, 140) { issue};
  \node[ipe node]
     at (180, 128) {7};
  \node[ipe node]
     at (200, 140) { today};
  \node[ipe node]
     at (208, 128) {8};
  \node[ipe node]
     at (232, 140) { .};
  \node[ipe node]
     at (232, 128) {9};
  \draw[-{ipe pointed[ipe arrow small]}]
    (48, 152.0004)
     arc[start angle=0, end angle=180, x radius=10.132, y radius=6.1918];
  \draw[-{ipe pointed[ipe arrow small]}]
    (76, 152.0004)
     arc[start angle=0, end angle=180, x radius=10, y radius=6.1111];
  \draw[-{ipe pointed[ipe arrow small]}]
    (52.0645, 152.0004)
     arc[start angle=0, end angle=180, x radius=-43.9675, y radius=26.869];
  \draw[-{ipe pointed[ipe arrow small]}]
    (84, 152.0004)
     arc[start angle=0, end angle=180, x radius=-12, y radius=7.3333];
  \draw[-{ipe pointed[ipe arrow small]}]
    (80.0003, 152)
     arc[start angle=0, end angle=180, x radius=-76.6917, y radius=46.8671];
  \draw[-{ipe pointed[ipe arrow small]}]
    (111.9995, 152.0004)
     arc[start angle=0, end angle=180, x radius=-50.1575, y radius=30.6518];
  \draw[-{ipe pointed[ipe arrow small]}]
    (143.9997, 152.0002)
     arc[start angle=0, end angle=180, x radius=-19.1473, y radius=19.5021];
  \draw[-{ipe pointed[ipe arrow small]}]
    (179.9997, 152.0003)
     arc[start angle=0, end angle=180, x radius=10.5407, y radius=9.6625];
  \node[ipe node, font=\tiny]
     at (28, 160) {ATT};
  \node[ipe node, font=\tiny]
     at (60, 160) {SBJ};
  \node[ipe node, font=\tiny]
     at (92, 160) {VC};
  \node[ipe node, font=\tiny]
     at (108, 180) {ATT};
  \node[ipe node, font=\tiny]
     at (157.096, 162.45) {ATT};
  \node[ipe node, font=\tiny]
     at (158.081, 173.023) {PC};
  \node[ipe node, font=\tiny]
     at (156, 184) {TMP};
  \node[ipe node, font=\tiny]
     at (152, 200) {PU};
  \draw[red]
    (124.623, 172.4461) circle[radius=2.2812];
  \draw[red]
    (93.8715, 178.667) circle[radius=2.7409];
\end{tikzpicture}
}
	\caption{An example of a sentence and the syntactic dependencies among its words (adapted from \newcite{Manual_TikzDep}). Relations are labeled with their grammatical category. Numbers below the sentence indicate the positions of the words. In this figure we see two syntactic crossings, marked with small red circles.}
	\label{fig:introduction:syntactic_dependency_structure:grammar}
\end{figure}

\begin{figure}
	\centering
\scalebox{0.9}{
\begin{tikzpicture}[ipe stylesheet]
  \node[ipe node]
     at (32, 544) {Yesterday};
  \node[ipe node]
     at (88, 544) {John};
  \node[ipe node]
     at (120, 544) {saw};
  \node[ipe node]
     at (140, 544) {a};
  \node[ipe node]
     at (152, 544) {dog};
  \node[ipe node]
     at (172, 544) {which};
  \node[ipe node]
     at (204, 544) {was};
  \node[ipe node]
     at (224, 544) {a};
  \node[ipe node]
     at (236, 544) {Yorkshire};
  \node[ipe node]
     at (280, 544) {Terrier};
  \draw[-{ipe pointed[ipe arrow small]}]
    (128, 552)
     arc[start angle=0, end angle=180, radius=38];
  \draw[-{ipe pointed[ipe arrow small]}]
    (128, 552)
     arc[start angle=0, end angle=180, radius=14];
  \draw[{ipe pointed[ipe arrow small]}-]
    (160, 552)
     arc[start angle=0, end angle=180, radius=16];
  \draw[-{ipe pointed[ipe arrow small]}]
    (158.399, 551.896)
     arc[start angle=-0.8277, end angle=180, radius=7.1998];
  \draw[{ipe pointed[ipe arrow small]}-]
    (212.0002, 552)
     arc[start angle=0, end angle=180.2424, radius=24.5802];
  \draw[-{ipe pointed[ipe arrow small]}]
    (210.2976, 551.934)
     arc[start angle=-0.2496, end angle=180, radius=15.1487];
  \draw[{ipe pointed[ipe arrow small]}-]
    (296.0002, 552)
     arc[start angle=0, end angle=179.3683, radius=40.5452];
  \draw[-{ipe pointed[ipe arrow small]}]
    (294.0719, 552.34)
     arc[start angle=0.5897, end angle=180, radius=33.0366];
  \draw[-{ipe pointed[ipe arrow small]}]
    (291.407, 552.562)
     arc[start angle=1.8187, end angle=180, radius=17.7079];
  \node[ipe node]
     at (104, 568) {1};
  \node[ipe node]
     at (144, 560) {1};
  \node[ipe node]
     at (184, 564) {1};
  \node[ipe node]
     at (260, 568) {1};
  \node[ipe node]
     at (240, 580) {2};
  \node[ipe node]
     at (180, 580) {2};
  \node[ipe node]
     at (80, 592) {2};
  \node[ipe node]
     at (232, 592) {3};
  \node[ipe node]
     at (48, 528) {1};
  \node[ipe node]
     at (96, 528) {2};
  \node[ipe node]
     at (124, 528) {3};
  \node[ipe node]
     at (140, 528) {4};
  \node[ipe node]
     at (156, 528) {5};
  \node[ipe node]
     at (180, 528) {6};
  \node[ipe node]
     at (208, 528) {7};
  \node[ipe node]
     at (224, 528) {8};
  \node[ipe node]
     at (252, 528) {9};
  \node[ipe node]
     at (292, 528) {10};
  \node[ipe node]
     at (32, 616) {a)};
  \node[ipe node]
     at (324, 712) {b)};
  \node[ipe node, font=\small]
     at (352, 664) {Yesterday};
  \node[ipe node, font=\small]
     at (400, 664) {John};
  \node[ipe node, font=\small]
     at (400, 716) {saw};
  \node[ipe node, font=\small]
     at (424, 624) {a};
  \node[ipe node, font=\small]
     at (432, 664) {dog};
  \node[ipe node, font=\small]
     at (424, 584) {which};
  \node[ipe node, font=\small]
     at (448, 624) {was};
  \node[ipe node, font=\small]
     at (452, 540) {a};
  \node[ipe node, font=\small]
     at (480, 540) {Yorkshire};
  \node[ipe node, font=\small]
     at (460, 584) {Terrier};
  \draw[-{ipe pointed[ipe arrow small]}]
    (408, 712)
     -- (376, 676);
  \draw[-{ipe pointed[ipe arrow small]}]
    (408, 712)
     -- (408, 676);
  \draw[-{ipe pointed[ipe arrow small]}]
    (408, 712)
     -- (440, 676);
  \draw[-{ipe pointed[ipe arrow small]}]
    (440, 656)
     -- (428, 632);
  \draw[-{ipe pointed[ipe arrow small]}]
    (440, 656)
     -- (456, 632);
  \draw[-{ipe pointed[ipe arrow small]}]
    (456, 620)
     -- (440, 596);
  \draw[-{ipe pointed[ipe arrow small]}]
    (456, 620)
     -- (476, 596);
  \draw[-{ipe pointed[ipe arrow small]}]
    (472, 580)
     -- (456, 552);
  \draw[-{ipe pointed[ipe arrow small]}]
    (472, 580)
     -- (496, 552);
  \node[ipe node]
     at (364, 652) {1};
  \node[ipe node]
     at (408, 652) {2};
  \node[ipe node]
     at (424, 716) {3};
  \node[ipe node]
     at (420, 612) {4};
  \node[ipe node]
     at (452, 664) {5};
  \node[ipe node]
     at (432, 572) {6};
  \node[ipe node]
     at (468, 624) {7};
  \node[ipe node]
     at (452, 528) {8};
  \node[ipe node]
     at (500, 528) {9};
  \node[ipe node]
     at (492, 584) {10};
\end{tikzpicture}
}
	\caption{a) An example of syntactic dependency structure. Arc labels indicate edge lengths, each calculated as the absolute difference of the positions of the corresponding edge's endpoints. The numbers below the words indicate positions. b) The rooted tree underlying the sentence in a); the positions of the words are indicated below or to the right of each word. Adapted from \cite[Figure 2]{McDonald2005a}.}
	\label{fig:introduction:syntactic_dependency_tree:lengths}
\end{figure}
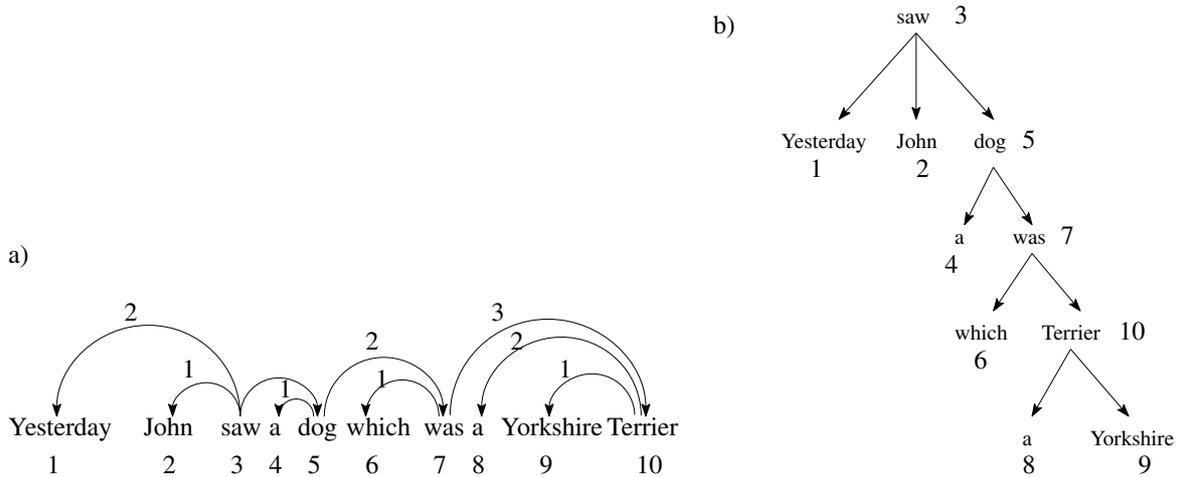

Quantitative Linguistics is a discipline within Linguistics that aims to unveil linguistic laws and explain their origins \cite{Kholer2012a,Best2017a}. Outstanding examples of these are Zipfian laws, e.g., Zipf's rank-frequency law, Zipf's law of abbreviation \cite{Zipf1949a}, that are defined typically on one of languages' basic units: words. Another discipline in Linguistics is Dependency Syntax, a framework which primarily reduces the syntactic structure of a sentence to word pairwise dependencies. Each of these dependencies has a `head' word and a `dependent' word (in fields like Computer Science, these could be called `parent' and `child', respectively; the `head' is also known as `governor'). The collection of such dependencies in a sentence combined with the linear ordering of the words yields the so-called {\em syntactic dependency structure} \cite{Melcuk1988a,Kuhlmann2006a,Nivre2006a,Gomez2011a} as in Figure \ref{fig:introduction:syntactic_dependency_structure:grammar}. Therefore, the underlying structure of a syntactic dependency structure can be seen as a rooted tree (as in Figure \ref{fig:introduction:syntactic_dependency_tree:lengths}(b)).

The combination of Quantitative Linguistics with Dependency Syntax has resulted into the emerging field of Quantitative Dependency Syntax \url{https://quasy-2019.webnode.com/}. The target of this field are syntactic dependency structures and aims to discover and understand statistical patterns in these structures. By linearizing the hierarchical structure ``arises the concept of dependency distance or dependency length'' \cite{Liu2017a}, defined usually as the number of intervening words between the endpoints of the dependency plus one \cite{Ferrer2004a} as in Figure \ref{fig:introduction:syntactic_dependency_tree:lengths}(a). Another relevant concept is that of {\em syntactic dependency crossing} \cite{Melcuk1988a}.
Figure \ref{fig:introduction:syntactic_dependency_structure:grammar} shows two examples of syntactic crossings: two syntactic dependencies cross when the positions of their head and dependent words interleave. Said concept is used to define many formal constraints, such as projective and planar structures \cite{Kuhlmann2006a} and 1-Endpoint-Crossing structures \cite{Pitler2013a}. See \newcite{Gomez2011a} for a review.

Research in Cognitive Science has shown a tendency for languages to reduce dependency distances \cite{Ferrer2004a,Liu2008a,Futrell2015a,Futrell2020a,Ferrer2022a}. According to \newcite{Liu2017a}, \newcite{Hudson1995a} gave the first definition of dependency distance and presented a cognitive formulation of ``the memory burden imposed by dependency distance on language processing''. This tendency results from the action of a Dependency Distance Minimization (DDm) principle \cite{Ferrer2003a,Ferrer2004a}, supported by many models and theories \cite{Liu2017a,Temperley2018a} stemming from the more general Principle of Least Effort \cite{Zipf1949a}, hence largely regarded as a linguistic universal. The statistical support for DDm comes from baselines that are used to perform statistical tests on the significance of dependency distances \cite{Ferrer2004a,Gildea2007a,Liu2008a,Park2009a,Gildea2010a,Futrell2015a,Yu2019a,Ferrer2022a}. Some of these baselines are defined on extreme conditions (e.g., maximum and minimum sum of dependency distances) which further motivates the study of extremal problems in Computer Science like the Minimum Linear Arrangement problem \cite{Garey1976a,Goldberg1976a,Shiloach1979a,Chung1984a} and the Maximum Linear Arrangement Problem \cite{Hassin2000a,Nurse2008a} and their variants under formal constraints. Other baselines are defined on `uniformly random' conditions, typically in uniformly random permutations of the words of a sentence. However, formal constraints from Dependency Grammar, like projectivity and planarity \cite{Sleator1993a,Kuhlmann2006a}, have led to defining such random baselines conditioned to those formal constraints \cite{Gildea2007a,Park2009a,Futrell2015a,Kramer2021a,Alemany2021a} although these formal constraints have been argued to be epiphenomena of DDm \cite{Gomez2016a,Gomez2019a}.

In this article we introduce a new tool to support research on the areas and the research problems reviewed above: the Linear Arrangement Library (LAL), which allows researchers to compute easily many of the metrics and algorithms that researchers have been proposing, while simplifying significantly the problem of calculating random or extremal baselines for them. In addition, LAL aims to simplify the process of working with collections of treebanks, one of the most successful recent examples being the Universal Dependencies collection \cite{UniversalDependencies26} and its variants \cite{SurfaceUniversalDependencies2018}. LAL is currently available from \url{https://cqllab.upc.edu/lal}.

In order to grasp the power of LAL we remind the reader that the syntactic dependency structure of a sentence can be defined as a triad composed of (1) a {\em directed graph structure} in which the vertices of the graph are the words of the sentence, (2) a {\em linear arrangement} of the vertices of the graph, and (3) {\em labels} of the edges of the graph which indicate the type of syntactic relationship between the words they relate. Two types of metrics (or scores) can be defined on such structures: word order-dependent, e.g., the sum of dependency distances \cite{Gildea2007a} and word order-independent metrics, e.g., mean hierarchical distance \cite{Jing2015a}. LAL allows one to compute many scores of each of these two sorts.

The calculation of baselines is at the heart of Quantitative Dependency Syntax research as well as at the heart of LAL. For some random baselines, LAL offers exact algorithms or formulae to calculate the desired value under the null model, e.g., algorithms to calculate the expected sum of dependency distances \cite{Ferrer2004a,Alemany2021a}. The reality is, unfortunately, that algorithms and formulae to calculate exact expected values of certain scores might be difficult to derive. As an alternative, LAL allows researchers to resort to random sampling in order to calculate said values, which often involves tree and/or linear arrangement generation \cite{Liu2008a,Esteban2017a,Yadav2019a}. An example of an application of tree generation would be the calculation of the expected mean hierarchical distance \cite{Jing2015a} among $n$-vertex rooted trees\footnote{This problem may be notoriously difficult to solve; take as a reference the work by \newcite{Renyi1965a} where it is shown that the average labeled tree height $H_n$ is such that $H_n\rightarrow \sqrt{2n\pi}$ as $n\rightarrow\infty$.}. Regarding linear arrangement generation, an example would be to calculate the expected flux weight \cite{Kahane2017a} over all arrangements of a tree.

Thanks to LAL the computation of random baselines can be restricted easily to two of the most frequently observed arrangements from a formal standpoint: planar orderings, where syntactic edges do not cross, and projective orders, namely planar orderings where the root is not covered \cite{Sleator1993a,Kuhlmann2006a}. For instance, the random baseline over the sum of dependency distances can be computed assuming unconstrained, projective and planar arrangements as shown in Table \ref{table:baselines}. In an unconstrained linear arrangement, edge crossings are allowed and the root may be covered.

The remainder of the article is organized as follows. Section \ref{sec:design_principles} presents the design principles of LAL and its architecture. Section \ref{sec:working_with_LAL} gives further details about its functionalities and explains how to work with LAL following a standard research pipeline. We end with some suggestions for future development of LAL. 

\begin{table}[]
	\centering
	\begin{tabular}{ccccc}
		\toprule
						& \multicolumn{3}{c}{$D$}									& $C$ \\
						\cmidrule{2-4}
						& Unconstrained		& Planar			& Projective		& Unconstrained \\
		\midrule
		Minimum			& \mymr{2}{3cm}{\newcite{Shiloach1979a}, \newcite{Chung1984a}}
											& \mymr{3}{3cm}{\newcite{Hochberg2003a}, \newcite{Alemany2022a}}
																& \mymr{3}{3cm}{\newcite{Gildea2007a}, \newcite{Alemany2022a}}
																					& $\dagger$ \\
						&					&					&					& \\
						&					&					&					& \\
						&					&					&					& \\
		Complexity		& $O(n^{2.2}),O(n^2)$& $O(n)$			& $O(n)$			& \\
		
		\midrule
		
		Expected		& \mymr{1}{3cm}{\newcite{Ferrer2004a}}
											& \mymr{1}{3cm}{}$^*$
																& \mymr{3}{3cm}{\newcite{Alemany2021a}}
																					& \mymr{1}{3cm}{\newcite{Verbitsky2008a}} \\
						&					&					&					& \\
						&					&					&					& \\
		Complexity		& $O(1)$			& $O(n)$			& $O(n)$			& $O(n)$ \\
		
		\midrule
		
		Maximum			& Under study		& In progress		& In progress		& Under study \\
		Complexity		&					& $O(n)$			& $O(n)$			& \\
		
		\bottomrule
	\end{tabular}
	\caption{The baselines on $D$, the sum of dependency distances, and $C$, the number of syntactic dependency crossings, that can be calculated on a given input tree using LAL. Columns `Unconstrained', `Planar' and `Projective' are the different constraints under which the `Minimum', `Expected' and `Maximum' values can be calculated with LAL. $^*$: available in LAL but article not published yet; $\dagger$: the minimum value of $C$ is trivially 0 for every tree.}
	\label{table:baselines}
\end{table}
\section{Design principles}
\label{sec:design_principles}

\subsection{Ease of use}

Many measures/scores in Quantitative Dependency Syntax have been devised over the years (e.g., \newcite{Jiang2018a}). Some of these are easy to calculate, e.g., the sum of dependency distances (or the sum of edge lengths) of an $n$-vertex syntactic dependency structure. This sum is easily computable in $O(n)$ time given the specification of the linear arrangement and that of the graph. However, the calculation of extremal values on linear arrangements (i.e. minimum or maximum values of a score), such as the solution to the Minimum Linear Arrangement (MLA) problem in unconstrained arrangements \cite{Garey1976a,Shiloach1979a,Chung1984a} or of one of its constrained variants \cite{Iordanskii1987a,Hochberg2003a,Gildea2007a,Bommasani2020a,Alemany2022a} are not straightforward because the algorithm is complex, it is hard to test or both. Likewise, performing statistical tests and calculating expected values (random baselines) require random sampling methods when exact algorithms/formulae are not known; such sampling is typically done uniformly at random over all possible trees \cite{Ferrer2018a,Gomez2019a,Yadav2019a} or random arrangements \cite{Ferrer2018a,Ferrer2022a}. LAL's main design principle is to make these algorithms (and random sampling methods) easily accessible and, therefore, LAL has been designed to be an easy-to-use tool for Quantitative Dependency Syntax researchers focused in the analysis of syntactic dependency trees, and Computer Scientists and Mathematicians specializing in Discrete Mathematics. Moreover, advanced programming knowledge is not required to use LAL. For example, a researcher in Quantitative Dependency Syntax can process a treebank as easily as shown in Code \ref{code:easy_example_use_LAL:treebank}.

\begin{code}
\begin{minted}{python}
import lal
err = lal.io.process_treebank("Cantonese.txt", "output_file.csv")
print(err)
\end{minted}
\caption{Python script for computing all scores on a single treebank with LAL. The input file is a series of head vectors, whose format is described in Section \ref{sec:working_with_LAL}, and the output is a standard .csv file that can be loaded directly on a spreadsheet.}
\label{code:easy_example_use_LAL:treebank}
\end{code}

\subsection{Connecting communities and traditions}

LAL aims to unite research traditions and disciplines from all over the world as well as serving the distinct fields converging into Quantitative Dependency Syntax as much as possible. LAL integrates views, concepts and scores from the major research communities: Asia (China, \newcite{Jing2015a}; Japan, \newcite{Komori2019a}), North America (USA, \newcite{Gildea2007a} and \newcite{Futrell2015a}) and Europe (e.g., France, \newcite{Kahane2017a}; Switzerland, \newcite{Gulordava2016a} and \newcite{Gulordava2015a}).

\subsection{Openness and availability}

After many years of research in Quantitative Dependency Syntax, the code used to calculate many of these metrics is not usually shared, thus forcing researchers in the Linguistics fields to repeat the same efforts as the original authors put into coding the algorithms. This repetition increases the probability of bugs in every researcher's code which lead to incorrect results used in their experiments to be published in scientific journals. We think that LAL is an answer to these challenges as it is an open-source project, licensed under the {\em GPL v3 Affero}\footnote{Many people think this license formally discourages commercial usage, but it certainly does not \url{https://www.gnu.org/licenses/agpl-3.0.en.html}.}. The library's code is publicly available at \linktogithub{linear-arrangement-library}.

\subsection{Robustness}

Often, testing is not thorough or is not specified. Therefore, the increasing amount of research in Quantitative Dependency Syntax creates a need for a thoroughly-tested tool in which to find many if not most of the metrics devised so far. The continual testing that is applied to LAL solves this problem: tests are run periodically to ensure that all algorithms calculate correct values.

\subsection{The architecture of LAL}

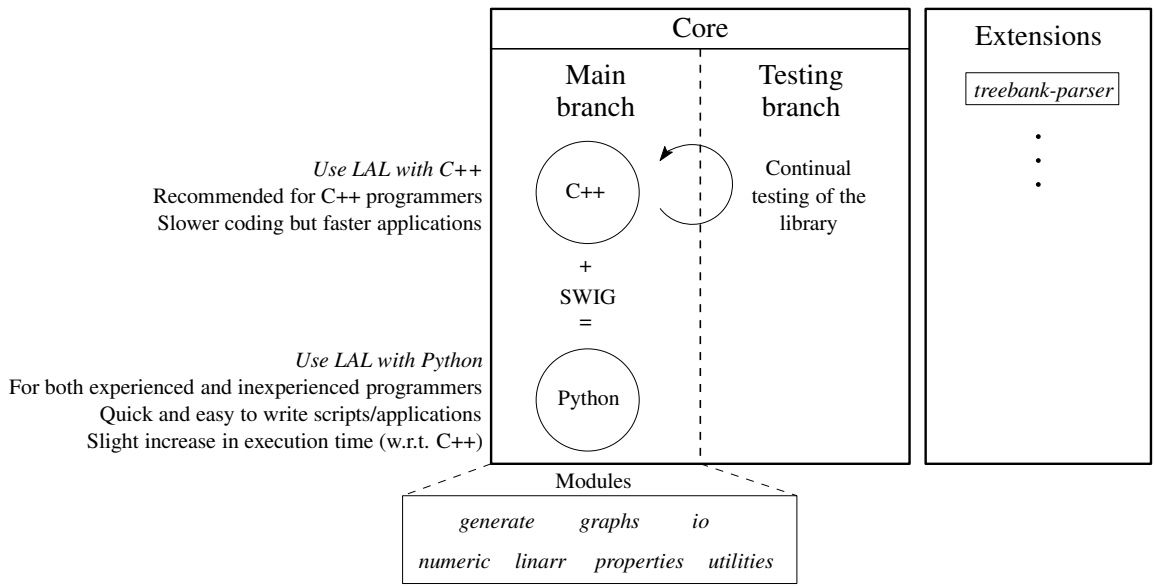
\begin{figure}
	\centering
\scalebox{0.75}{
\begin{tikzpicture}[ipe stylesheet]
  \draw[ipe pen fat]
    (248, 788) rectangle (456, 560);
  \node[ipe node, anchor=north west]
     at (376, 760) {
       \begin{minipage}{52bp}\Large\kern0pt
         \begin{center}
         Testing branch
         \end{center}
       \end{minipage}
     };
  \node[ipe node, anchor=north west]
     at (272, 760) {
       \begin{minipage}{56bp}\Large\kern0pt
         \begin{center}
         Main branch
         \end{center}
       \end{minipage}
     };
  \node[ipe node, anchor=north west]
     at (376, 712) {
       \begin{minipage}{60bp}\kern0pt
         \begin{center}
         Continual testing of the library
         \end{center}
       \end{minipage}
     };
  \draw[-ipe pointed]
    (332, 688)
     arc[start angle=36.8699, end angle=323.1301, radius=-20];
  \node[ipe node]
     at (285.152, 693.316) {C++};
  \draw
    (296, 696) circle[radius=25.6125];
  \draw[ipe pen heavier, ipe dash dashed]
    (352, 768)
     -- (352, 560);
  \node[ipe node]
     at (281.152, 589.316) {Python};
  \draw
    (296, 592) circle[radius=25.6125];
  \node[ipe node]
     at (282.239, 640) {SWIG};
  \node[ipe node, anchor=north west]
     at (76, 712) {
       \begin{minipage}{168bp}\kern0pt
         \begin{flushright}
         {\em Use LAL with C++}\\
         Recommended for C++ programmers\\
         Slower coding but faster applications
         \end{flushright}
       \end{minipage}
     };
  \node[ipe node, anchor=north west]
     at (8, 616) {
       \begin{minipage}{236bp}\kern0pt
         \begin{flushright}
         {\em Use LAL with Python}\\
         For both experienced and inexperienced programmers\\
         Quick and easy to write scripts/applications\\
         Slight increase in execution time (w.r.t. C++)
         \end{flushright}
       \end{minipage}
     };
  \node[ipe node]
     at (232.191, 528.025) {{\em generate}};
  \node[ipe node]
     at (292, 528) {{\em graphs}};
  \node[ipe node]
     at (348.024, 527.668) {{\em io}};
  \node[ipe node]
     at (212.119, 507.835) {{\em numeric}};
  \node[ipe node]
     at (260, 508) {{\em linarr}};
  \node[ipe node]
     at (299.857, 508.001) {{\em properties}};
  \node[ipe node]
     at (356, 508) {{\em utilities}};
  \draw
    (204, 544) rectangle (400, 500);
  \draw[ipe dash dashed]
    (248, 560)
     -- (204, 544);
  \draw[ipe dash dashed]
    (352, 560)
     -- (400, 544);
  \node[ipe node]
     at (280, 548) {Modules};
  \node[ipe node]
     at (292, 656) {+};
  \node[ipe node]
     at (292, 628) {=};
  \draw[ipe pen fat]
    (464, 788) rectangle (576, 560);
  \node[ipe node, anchor=north west]
     at (476, 780) {
       \begin{minipage}{88bp}\Large\kern0pt
         \begin{center}
         Extensions
         \end{center}
       \end{minipage}
     };
  \node[ipe node]
     at (488.191, 744.025) {{\em treebank-parser}};
  \draw
    (484, 756) rectangle (560, 740);
  \pic[ipe mark small]
     at (520, 724) {ipe disk};
  \pic[ipe mark small]
     at (520, 712) {ipe disk};
  \pic[ipe mark small]
     at (520, 700) {ipe disk};
  \draw[ipe pen heavier]
    (248, 768)
     -- (456, 768);
  \node[ipe node, anchor=north west]
     at (264, 784) {
       \begin{minipage}{176bp}\Large\kern0pt
         \begin{center}
         Core
         \end{center}
       \end{minipage}
     };
\end{tikzpicture}
}
	\caption{The architecture of LAL project: {\em core} and {\em extensions}. The core consists of the library and testing branch. The library is developed in C++, and wrapped into Python via \newcite{Web_SWIG}. Testing the library is a crucial part of its development. The library is composed of 7 modules (both C++ and Python options): \module{generate}, \module{graphs}, \module{io}, \module{numeric}, \module{linarr}, \module{properties}, and \module{utilities}, each of which is briefly described in Section \ref{sec:design_principles}.}
	\label{fig:core_design}
\end{figure}

The LAL project's architecture consists of a {\em core} and {\em extensions} (Figure \ref{fig:core_design}). The core has two parts: the main branch and the testing branch (Figure \ref{fig:core_design}). The latter is responsible for the robustness of the main branch. The test branch is not publicly available yet but it results from the transfer of knowledge and methods that enabled to test and correct classic algorithms with random and exhaustive methods \cite{Esteban2017a,Alemany2020a,Alemany2022a}. The C++ language is used in LAL's core to implement its main functionalities and algorithms, some of which are parallelized internally to improve performance (a critical example is the function that computes (all) scores on a treebank collection). However, only the library branch is wrapped to Python (via \newcite{Web_SWIG}; Figure \ref{fig:core_design}) given the fact that Python is usually the first choice of many scientists who are looking forward to automatizing their workflow as it is easy to use.
 
LAL extensions implement additional functionalities, e.g., dealing with the interface between existing treebanks and LAL. `LAL extensions' is the place for contributing to the LAL ecosystem without knowledge on how LAL is implemented. A concrete LAL extension is introduced in the next section.

LAL's core is composed of 7 modules (Figure \ref{fig:core_design}). Now follows a brief description of each of them.
\begin{itemize}
	\item The \module{generate} module contains algorithms to generate trees uniformly at random or exhaustively, labeled or unlabeled, and free or rooted; algorithms to generate arrangements of trees under the projectivity or planarity constraint,

	\item In the \module{graphs} module users will find the implementations of different graph classes: undirected and directed graphs, free and rooted trees,

	\item The \module{io} (input/output) module contains algorithms to read data from a file in disk, but its current chief goal is to process input data, such as treebanks, to produce an output file with measures computable by the library,

	\item The \module{numeric} module contains wrappers of the GMP library \cite{Web_GMP} for arbitrary-precision integer and rational numbers to ensure exact precision in the calculations (mostly useful for the testing branch of the project or advanced research in mathematics),

	\item In the \module{linarr} module users will find many algorithms to compute measures of graphs that are defined on the linear ordering of the vertices, e.g., the sum of dependency distances \cite{Gildea2007a},

	\item The \module{properties} module implements algorithms to compute measures of graphs that do not depend on the linear ordering of their vertices, only on their structure, e.g., the Mean Hierarchical Distance \cite{Jing2015a},

	\item Finally, the \module{utilities} module contains algorithms that are interesting to the general public but cannot be classified into the categories above, such as the tree isomorphism test that tells whether or not two trees are the same. This test only takes into account the structure of the tree (nodes and edges) and not possible labels on the edges (e.g. grammatical relations between words in a sentence) or possible tokens in the nodes (e.g. words from a sentence). The algorithm implemented in LAL \cite{Aho1974a} is an $O(n)$-time algorithm (where n is the number of vertices of the trees) adapted to the case when the two trees are free or rooted.

\end{itemize}

\section{Working with LAL}
\label{sec:working_with_LAL}

LAL's capabilities range over three kinds of operations depending on the entity to which they are applied: operations on {\em individual} syntactic dependency structures (modules \module{graphs}, \module{linarr}, \module{properties}), operations on individual {\em treebanks} or on treebank {\em collections} (\module{io} module).

LAL contains many functions with which one can calculate metrics/scores on a single tree. At present, the focus of LAL is on trees for simplicity and because of the high current interest on this kind of graphs. However, although fewer in number, LAL also contains functions that admit general graphs (e.g., graphs with cycles, and forests). Functions might depend on a given word ordering (\module{linarr} module) or not (\module{properties} module). Most functions in the \module{linarr} module typically require a linear arrangement as an input parameter. However, some functions return a linear arrangement rather than requiring one as an input parameter, as in the example given in Code \ref{code:easy_example_use_LAL:Dmin}. It is worth mentioning that LAL implements methods for a flexible construction of graphs (\module{graphs} module) by adding edges one by one, or in bulk, thus allowing complicated workflows, whenever those are needed. Nevertheless, LAL is also equipped with helper functions for simpler graph construction (e.g., construct a graph directly from its set of edges). Furthermore, it provides functions for reading graphs from a file in its \module{io} module.

\begin{code}
\begin{minted}{python}
import lal
t = lal.graphs.from_edge_list_to_free_tree([(0,1), ...])
Shiloachs_algorithm = lal.linarr.algorithms_Dmin.Shiloach
print(lal.linarr.min_sum_edge_lengths(t), Shiloachs_algorithm)
Chungs_algorithm = lal.linarr.algorithms_Dmin.Chung_2
print(lal.linarr.min_sum_edge_lengths(t), Chungs_algorithm)
\end{minted}
\caption{Python script to calculate the minimum baseline for the sum of dependency distances on a free tree with LAL applying two different algorithms: Shiloach's \cite{Shiloach1979a,Esteban2017a} and Chung's quadratic algorithm \cite{Chung1984a}. LAL implements the correction of Shiloach's algorithm in \cite{Esteban2017a}. The free tree is specified as a list of pairs of edges. That baseline is the solution of the so-called minimum linear arrangement problem of computer science, hence the names of the algorithms mentioned above.}
\label{code:easy_example_use_LAL:Dmin}
\end{code}

The library also implements treebank processing (\module{io} module). It provides its users with algorithms to generate data out of single treebank files and collections of treebanks. A collection of treebanks is simply a set of treebanks that are related to one another within some particular context. In the UD collection, texts from distinct languages annotated with the same formalism \cite{UniversalDependencies26}, but there are many other possibilities: e.g., the novels of the same writer -- where an individual treebank is a novel, all the articles in a given newspaper -- where a single treebank is one of said articles.

In LAL a collection is represented by a plain text file listing all the treebank files in the collection.
Treebanks are typically written in CoNLL-U format \cite{Buchholz2006a}, but LAL requires a simpler format with the essential information. In particular, LAL requires treebanks to be provided as a series of {\em head vectors} (or {\em head sequences}); a head vector of an $n$-word sentence is a sequence of $n$ non-negative integer numbers in the positions from $1$ to $n$ where each number indicates the position of its parent word, and the number $0$ indicates the root word of the sentence. The most important reason to use the intermediate format is the existence of many previous formats, and the possibility of many new formats coming to life; we believe it will help in reducing the library's usage complexity. This approach is similar to that taken by compiler developers and designers who use the so-called `Three address code': it is an intermediate language into which many programming languages can be transformed, and is then compiled to the target machine. Nevertheless, we have also developed a LAL extension (Figure \ref{fig:core_design}), the {\em treebank-parser} that applies core LAL functions to convert CoNLL-U-formatted files into the head vector format after applying optional preprocessing: (a) removal of punctuation marks, e.g., for crosslinguistic generality, (b) removal of functions words, e.g., to compare languages with many such words against languages where these are scarce and (c) removal of sentences shorter or longer than a given length \cite{Ferrer2022a}. This tool can be found online at \linktogithub{treebank-parser}.

The processing of treebanks (and collection of treebanks) can be done automatically by LAL, but it can also be customized by its users. In other words, the automatic processing of a treebank (or a collection of treebanks) is performed by the library applying the metrics/scores selected by the user, with optional internal parallelization, and other customizable options on the format of the output. One obvious drawback of automatic processing is that it is limited to the metrics/scores implemented in LAL. Nevertheless, should users look forward to calculating a new score on a treebank (or in a collection), they can still use LAL to carry out such a task since LAL implements algorithms to easily iterate over a file containing only head vectors, thus removing the aforementioned limitation. Therefore, LAL facilitates working on a single treebank or on a collection of treebanks.

\begin{figure}
	\centering
\begin{tikzpicture}[ipe stylesheet]
  \node[ipe node, anchor=north west]
     at (212, 772) {
       \begin{minipage}{96bp}\kern0pt
         \begin{center}
         Handwritten/Typeset text
         \end{center}
       \end{minipage}
     };
  \node[ipe node, anchor=north west]
     at (240, 716) {
       \begin{minipage}{44bp}\kern0pt
         PARSER
       \end{minipage}
     };
  \node[ipe node, anchor=north west]
     at (232, 672) {
       \begin{minipage}{60bp}\kern0pt
         TREEBANK
       \end{minipage}
     };
  \draw[ipe pen heavier]
    (232, 724) rectangle (288, 700);
  \draw[shift={(217.673, 750.666)}, xscale=0.8821, yscale=0.8333]
    (0, 0)
     .. controls (24, -8) and (72, -8) .. (96, 0)
     .. controls (120, 8) and (120, 24) .. (96, 32)
     .. controls (72, 40) and (24, 40) .. (0, 32)
     .. controls (-24, 24) and (-24, 8) .. cycle;
  \draw[shift={(233.333, 653.334)}, yscale=0.6667]
    (0, 0)
     .. controls (13.3333, -8) and (40, -8) .. (53.3333, 0)
     .. controls (66.6667, 8) and (66.6667, 24) .. (53.3333, 32)
     .. controls (40, 40) and (13.3333, 40) .. (0, 32)
     .. controls (-13.3333, 24) and (-13.3333, 8) .. cycle;
  \node[ipe node, anchor=north west]
     at (336, 675.122) {
       \begin{minipage}{72bp}\kern0pt
         \begin{center}
         UD, Prague, Stanford, $\cdots$
         \end{center}
       \end{minipage}
     };
  \draw[shift={(340, 653.489)}, yscale=0.7273]
    (0, 0)
     .. controls (16, -7.5458) and (48, -7.5458) .. (64, -0.2125)
     .. controls (80, 7.1208) and (80, 21.7875) .. (64, 29.3333)
     .. controls (48, 36.8792) and (16, 37.3042) .. (0, 29.9708)
     .. controls (-16, 22.6375) and (-16, 7.5458) .. cycle;
  \node[ipe node, anchor=north west]
     at (276, 572) {
       \begin{minipage}{32bp}\kern0pt
         PREPROCESSING
       \end{minipage}
     };
  \draw[ipe pen heavier]
    (268, 580) rectangle (368, 556);
  \draw
    (316, 632)
     -- (300, 616)
     -- (316, 600)
     -- (332, 616)
     -- cycle;
  \draw[-ipe pointed]
    (260, 648)
     -- (308, 624);
  \draw[-ipe pointed]
    (372, 648)
     -- (324, 624);
  \node[ipe node]
     at (224, 612) {choice of source};
  \draw[-ipe pointed]
    (264, 559.3286)
     arc[start angle=29.7449, end angle=330.2551, x radius=15.7703, y radius=-15.7703];
  \node[ipe node, anchor=north west]
     at (124, 584) {
       \begin{minipage}{108bp}\kern0pt
         \begin{center}
         Removal of punctuation marks, perhaps function words, ...
         \end{center}
       \end{minipage}
     };
  \node[ipe node]
     at (254.659, 656) {(1)};
  \node[ipe node, anchor=north west]
     at (288, 528) {
       \begin{minipage}{60bp}\kern0pt
         TREEBANK
       \end{minipage}
     };
  \draw[shift={(289.333, 509.334)}, yscale=0.6667]
    (0, 0)
     .. controls (13.3333, -8) and (40, -8) .. (53.3333, 0)
     .. controls (66.6667, 8) and (66.6667, 24) .. (53.3333, 32)
     .. controls (40, 40) and (13.3333, 40) .. (0, 32)
     .. controls (-13.3333, 24) and (-13.3333, 8) .. cycle;
  \node[ipe node]
     at (310.026, 512) {(2)};
  \node[ipe node, anchor=north west]
     at (160, 532) {
       \begin{minipage}{108bp}\kern0pt
         \begin{center}
         Processed collection of syntactic dependency trees
         \end{center}
       \end{minipage}
     };
  \draw[-ipe pointed]
    (260, 700)
     -- (260, 680);
  \node[ipe node, anchor=north west]
     at (272, 476) {
       \begin{minipage}{88bp}\kern0pt
         \begin{center}
         HEAD VECTOR GENERATOR
         \end{center}
       \end{minipage}
     };
  \draw[ipe pen heavier]
    (268, 484) rectangle (364, 448);
  \node[ipe node, anchor=north west]
     at (288, 420) {
       \begin{minipage}{60bp}\kern0pt
         TREEBANK
       \end{minipage}
     };
  \draw[shift={(289.333, 401.334)}, yscale=0.6667]
    (0, 0)
     .. controls (13.3333, -8) and (40, -8) .. (53.3333, 0)
     .. controls (66.6667, 8) and (66.6667, 24) .. (53.3333, 32)
     .. controls (40, 40) and (13.3333, 40) .. (0, 32)
     .. controls (-13.3333, 24) and (-13.3333, 8) .. cycle;
  \node[ipe node]
     at (310.52, 404) {(3)};
  \node[ipe node, anchor=north west]
     at (112, 680) {
       \begin{minipage}{108bp}\kern0pt
         \begin{center}
         Non-processed collection of syntactic dependency trees
         \end{center}
       \end{minipage}
     };
  \node[ipe node, anchor=north west]
     at (208, 420) {
       \begin{minipage}{64bp}\kern0pt
         \begin{center}
         Collection of head vectors
         \end{center}
       \end{minipage}
     };
  \filldraw[lightblue, ipe pen fat]
    (244, 376) rectangle (388, 352);
  \node[ipe node, anchor=north west]
     at (360, 436) {
       \begin{minipage}{104bp}\kern0pt
         0 1 2 6 4 1 6 6 6\\
         0 1 2 5 1 7 1 7 10 7 10\\
         2 0 2 2 4 4 8 4 8 9\\
         $\cdots$
       \end{minipage}
     };
  \node[ipe node, anchor=north west]
     at (140, 720) {
       \begin{minipage}{80bp}\kern0pt
         \begin{center}
         parsing algorithm or a human
         \end{center}
       \end{minipage}
     };
  \draw[-ipe pointed]
    (260, 744)
     -- (260, 724);
  \draw[-ipe pointed]
    (316, 600)
     -- (316, 580);
  \draw[-ipe pointed]
    (316, 556)
     -- (316, 536);
  \draw[-ipe pointed]
    (316, 504)
     -- (316, 484);
  \draw[-ipe pointed]
    (316, 448)
     -- (316, 428);
  \draw[-ipe pointed]
    (316, 396)
     -- (316, 376);
  \node[ipe node]
     at (324, 688) {Existing TREEBANK};
  \node[ipe node, anchor=north west]
     at (207.792, 272.14) {
       \begin{minipage}{128bp}\kern0pt
         \begin{center}
         Hypothesis and theoretical prediction testing
         \end{center}
       \end{minipage}
     };
  \draw[-ipe pointed]
    (316, 352)
     -- (316, 332);
  \node[ipe node, anchor=north west]
     at (252, 368) {
       \begin{minipage}{128bp}\kern0pt
         \begin{center}
         Linear Arrangement Library
         \end{center}
       \end{minipage}
     };
  \node[ipe node]
     at (307.643, 319.643) {.csv};
  \draw
    (303.607, 315.1548)
     .. controls (309.607, 311.8214) and (321.607, 311.8214) .. (327.607, 315.1548)
     .. controls (333.607, 318.4881) and (333.607, 325.1548) .. (327.607, 328.4881)
     .. controls (321.607, 331.8214) and (309.607, 331.8214) .. (303.607, 328.4881)
     .. controls (297.607, 325.1548) and (297.607, 318.4881) .. cycle;
  \draw
    (202.153, 292.084) rectangle (430.153, 244.084);
  \node[ipe node, anchor=base]
     at (314.331, 280.309) {{\bf STATISTICAL ANALYSIS}};
  \draw[-ipe pointed]
    (316, 312)
     -- (316, 292);
  \draw[ipe dash dashed]
    (63.927, 590.0445)
     -- (499.927, 590.0445);
  \draw[ipe dash dashed]
    (63.927, 494.0445)
     -- (499.927, 494.0445);
  \draw[ipe dash dashed]
    (63.927, 386.0445)
     -- (499.927, 386.0445);
  \node[ipe node, font=\LARGE]
     at (64, 660) {\rotatebox{90}{Phase 1}};
  \node[ipe node, font=\LARGE]
     at (64, 516) {\rotatebox{90}{Phase 2}};
  \node[ipe node, font=\LARGE]
     at (64, 416) {\rotatebox{90}{Phase 3}};
  \node[ipe node, font=\LARGE]
     at (64, 316) {\rotatebox{90}{Phase 4}};
  \node[ipe node, anchor=north west]
     at (343.792, 272.14) {
       \begin{minipage}{80bp}\kern0pt
         \begin{center}
         Evaluation of a treebank's quality
         \end{center}
       \end{minipage}
     };
  \draw[ipe dash dashed]
    (63.927, 302.0445)
     -- (499.927, 302.0445);
  \node[ipe node, font=\LARGE]
     at (64, 244) {\rotatebox{90}{Phase 5}};
  \node[ipe node, anchor=north west]
     at (376, 584) {
       \begin{minipage}{108bp}\kern0pt
         \begin{center}
         See LAL extensions (Figure \ref{fig:core_design}) for a possible tool to carry this out
         \end{center}
       \end{minipage}
     };
\end{tikzpicture}
	\caption{A standard pipeline for research on a single treebank. The pipeline comprises five phases that start with the choice of an existing treebank or producing it from raw text (Phase 1) and ends with analyses of the output produced by LAL (Phase 5). In Phase 4, LAL receives a treebank that has been preprocessed and transformed into a format that LAL can digest.}
	\label{fig:pipeline_process_treebank}
\end{figure}
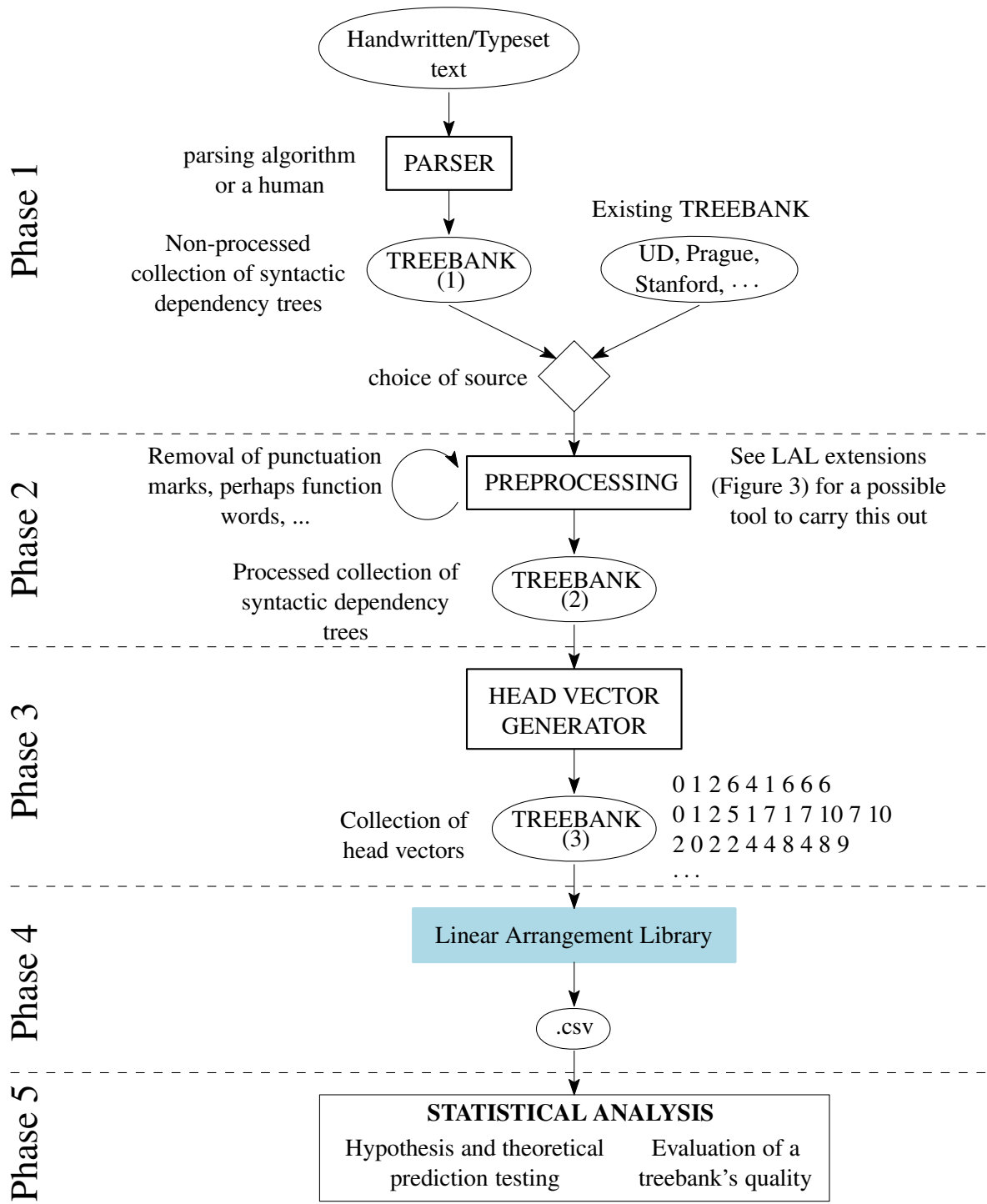

Figure \ref{fig:pipeline_process_treebank} illustrates one possible pipeline when working with a single treebank. That pipeline can easily be adapted to any treebank collection. A description now follows.
\begin{itemize}
\item[Phase 1] Firstly, one chooses a source of the input data. Such data may come from a handwritten text, or a typeset text, which is to be analyzed by either a computer or by a human; the data may come from already-analyzed data such as the UD collection \cite{UniversalDependencies26,SurfaceUniversalDependencies2018}, or treebanks annotated with the Stanford \cite{StanfordDependencies} or Prague \cite{PragueDependencies} conventions.

\item[Phase 2] Secondly, some preprocessing of the data might be required, e.g., removal of punctuation marks, removal of function words, or the exclusion of sentences that are too short or too long. See \newcite{Ferrer2022a} for a complete example of such preprocessing. We call the resulting treebank `Treebank (2)'.

\item[Phase 3] Then, one transforms `Treebank (2)' into `Treebank (3)', a file containing only the so-called head vectors so that LAL can understand the data.

\item[Phase 4] Here, LAL is used to generate a {\tt .csv} file containing all the measures that are interesting for one's own research. Staying true to its efficiency design principle, LAL can evaluate all metrics it implements on every tree of the UD 2.6 treebank \cite{UniversalDependencies26} in $\sim 23$ seconds, while producing only the primary data (on only the UD 2.6 treebank) of a recent article \cite{Ferrer2022a} takes $\sim 6$ seconds\footnote{Running times are estimated on a brand-new PC with a 3.30 GHz, 6-core (2 threads/core) {\em i5-10600} CPU (16 GB); the program uses 6 threads and runs on Ubuntu 20.04.}. 

\item[Phase 5] The last phase consists of using the data to perform statistical analyses on the treebank. These analyses comprise hypothesis and theoretical prediction testing \cite{Ferrer2019b} as well as evaluation of the quality of a treebank \cite{Alzetta2017a,Heinecke2019a}. When the pipeline is adapted to a treebank collection, LAL supports research in typology \cite{Croft2017a,Alzetta2018a} or on comparisons of annotation schemes \cite{Osborne2019a,Passarotti2016a}.
\end{itemize}

LAL extensions can be used for tasks in Phases 2 and 3. For the time being, the aforementioned extension {\em treebank-parser} (Figure \ref{fig:core_design}) allows one to perform Phase 2 and Phase 3 in a row over treebanks in CoNLL-U format. In the future, we hope that LAL extensions grow in number with contributions that researchers wish to make to the LAL ecosystem for Phase 2 or Phase 3.

\subsection{Capabilities}

To begin with, LAL implements several algorithms to calculate relevant word order-dependent metrics (module \module{linarr}) in Quantitative Dependency Syntax. These include the sum of dependency distances \cite{Gildea2007a}; the number of edge crossings \cite{Ferrer2018a}, with the option to choose among several algorithms; the prediction of the number of crossings based on the length of the edges \cite{Ferrer2014a}; the class of syntactic dependency structure (such as WG$_1$ \cite{Gomez2011a}, 1-Endpoint Crossing \cite{Pitler2013a}, projective and planar \cite{Sleator1993a,Kuhlmann2006a}); the solution to the MLA problem under several constraints (Table \ref{table:baselines}); the computation of dependency fluxes \cite{Kahane2017a}; the proportion of head initial dependencies \cite{Liu2010a}.

Researchers will find in LAL many word order-independent metrics (metrics that do not depend on the ordering of the vertices) in the \module{properties} module, including the Mean Hierarchical Distance \cite{Jing2015a}; the variance and expected number of crossings in unconstrained arrangements of trees, useful for variable standardization \cite{Alemany2020a}; the expected sum of edge lengths under three different formal constraints (Table \ref{table:baselines}) to cover different views about the nature of formal constraints \cite{Yadav2021a}, and the variance of said sum in unconstrained linear arrangements \cite{Ferrer2019a}; the calculation of the $m$-th moment of degrees about zero used to calculate the well-known hubiness coefficient \cite{Ferrer2018a}, extended to the out- and in-degrees\footnote{ Trivially, the $m$-th moment of in-degree about zero of any $n$-vertex tree is $\langle k_{in}^m \rangle = (n-1)/n$. Nevertheless, a function to calculate it is provided.}; the number of pairs of independent edges (two edges are independent when they share no vertices); the central and centroidal vertices of trees \cite{Harary1969a}. Furthermore, one can classify trees into classes according to their structure. These classes are: {\em linear}, {\em star}, {\em quasistar} and {\em bistar} \cite{SanDiego2014a}, {\em caterpillar} \cite{Harary1973a}, and {\em spider} \cite{Bennett2019a} trees.

In order to overcome the research limitations arising from the lack of research on every possible expected value that can be elicited, LAL is equipped with several algorithms for the generation of trees (Table \ref{table:tree_generation}) and of linear arrangements of trees under several formal constraints (projectivity and planarity\footnote{ Random and exhaustive generation of planar arrangements is available in LAL but not published at the time of submission.} \cite{Kuhlmann2006a}), and, for the sake of ease of use, it also provides exhaustive and random generation of unconstrained arrangements, i.e., exhaustive and random generation of permutations. Random generation allows one to estimate, via random sampling of the appropriate structure, the expected value of a certain existing or new metric. For example, one could easily approximate the expected Mean Hierarchical Distance (MHD) over the set of uniformly random unlabeled rooted trees by sampling said trees and averaging their MHD. The same can be said about those metrics dependent on word order. It goes without saying that said approximation is not limited to what LAL can calculate: researchers with sufficient programming skills can implement their own metrics either in C++ or in Python and approximate that metric's expectation and other aspects of its distribution under null models; the online documentation at \url{https://cqllab.upc.edu/lal/guides} explains how these estimates can be calculated and provide examples that inexperienced programmers can easily adapt to the metric of their choice. In previous research, the methods of generation of random trees do not warrant uniform sampling of the space of possible trees \cite{Liu2007a,Yan2019a}. LAL simplifies research where uniformity is required.

\begin{table}[]
	\centering
	\begin{tabular}{cccc}
		\toprule
		Sampling method & \multicolumn{2}{c}{Type of tree} & References \\
		\midrule
		\multirow{4}{*}{Exhaustive}
			& \multirow{2}{*}{Labeled}		& Free		& \newcite{Pruefer1918a} \\
			&								& Rooted	& Based on E-L-F$^*$ \\
			\cmidrule{2-4}
			& \multirow{2}{*}{Unlabeled}	& Free		& \newcite{Wright1986a} \\
			&								& Rooted	& \newcite{Beyer1980a} \\
		\midrule
		\multirow{4}{*}{Random}
			& \multirow{2}{*}{Labeled}		& Free		& \newcite{Pruefer1918a} \\
			&								& Rooted	& Based on Rn-L-F${^*}{^*}$ \\
			\cmidrule{2-4}
			& \multirow{2}{*}{Unlabeled}	& Free		& \newcite{Wilf1981a}${^*}{^*}{^*}$ \\
			&								& Rooted	& \newcite{Nijenhuis1978a} \\
		\bottomrule
	\end{tabular}
	\caption{The eight different possibilities of tree generation in LAL. By `random' we mean `uniformly random over the complete set of the respective kind of trees'. $^*$ Based on Exhaustive-Labeled-Free tree generation. ${^*}{^*}$ Based on Random-Labeled-Free tree generation. ${^*}{^*}{^*}$ The algorithm includes the correction pointed out in \cite{GiacXcas_Manual}.}
	\label{table:tree_generation}
\end{table}

\subsection{Applications of LAL}

LAL's domain of application revolves primarily around studies on Quantitative Dependency Syntax. 
We have dealt with various applications of LAL when describing Phase 5 of the pipeline (Section \ref{sec:working_with_LAL}, Figure \ref{fig:pipeline_process_treebank}). LAL is also a convenient tool for reproducing results from previous research. For instance, LAL allows one to reproduce the results of the analyses to predict the actual number of dependency crossings \cite{Gomez2016a}, the results on the scaling of sum of dependency distances in minimum linear arrangements \cite{Esteban2016a}, or recent findings on the degree of optimality of languages \cite{Ferrer2022a}. This research eventually converged into the formal development of LAL.

LAL offers many possibilities for research on the similarity among trees. For instance, LAL can be used to find how many isomorphic tree structures are present in two treebanks. Also, given any treebank, one can also find the amount of unique trees (up to graph isomorphism). Furthermore, rooted trees are implemented so that users can extract rooted subtrees and perform similarity tests based on subgraph isomorphism. These have already been tackled for general graphs \cite{Cordella2004a,Juttner2018a}.

Beyond the realm of measurements on treebanks, LAL contains tree-generation (or linear-arrangement generation) methods that can help one to test new algorithms. With this one can easily implement the testing protocol to assert the correctness of the implementation of the algorithm to compute the minimum sum of dependency distances \cite{Esteban2016a,Esteban2017a}.

\section{Future work}

LAL is a growing project to support current and future research. We plan to extend the library with algorithms to calculate extreme or random baselines for scores for which a fast exact algorithm is not available yet. For instance, we plan to add efficient algorithms for the calculation of the maximum sum of edge lengths in unconstrained arrangements and also on said maximum under the projectivity and planarity constraint (Table \ref{table:baselines}). We will also work on the classification of linear arrangements into different classes of formal constraints and also extend the library with functionalities that ease common tasks in the analysis of syntactic dependency structures or that reflect consolidated results from the distinct disciplines involved. We are also open to update the library based on demands of engaged users.

\section*{Acknowledgements}

We thank Aleksandra Petrova for helpful comments. LAP is supported by Secretaria d’Universitats i Recerca de la Generalitat de Catalunya and the Social European Fund. RFC and LAP are supported by the grant TIN2017-89244-R from MINECO (Ministerio de Econom\'{i}a, Industria y Competitividad). RFC is also supported by the recognition 2017SGR-856 (MACDA) from AGAUR (Generalitat de Catalunya). JLE is funded by the grant PID2019-109137GB-C22 from MINECO.

\clearpage
\bibliographystyle{acl}

\begin{thebibliography}{}

\bibitem[\protect\citename{Aho \bgroup et al.\egroup }1974]{Aho1974a}
{Alfred V.} Aho, {Jeffrey E.} Hopcroft, and {John D.} Ullman.
\newblock 1974.
\newblock {\em The {D}esign and {A}nalysis of {C}omputer {A}lgorithms}.
\newblock Addison-Wesley series in computer science and information processing.
  Addison-Wesley Publishing Company, Michigan University, 1st edition.

\bibitem[\protect\citename{{Alemany-Puig} and
  {Ferrer-i-Cancho}}2020]{Alemany2020a}
{Llu\'{i}s} {Alemany-Puig} and Ramon {Ferrer-i-Cancho}.
\newblock 2020.
\newblock Edge crossings in random linear arrangements.
\newblock {\em Journal of Statistichal Mechanics}, 2020:023403.

\bibitem[\protect\citename{{Alemany-Puig} and
  {Ferrer-i-Cancho}}2021]{Alemany2021a}
Llu\'{i}s {Alemany-Puig} and Ramon {Ferrer-i-Cancho}.
\newblock 2021.
\newblock Linear-time calculation of the expected sum of edge lengths in
  projective linearizations of trees.
\newblock {\em arXiv}.

\bibitem[\protect\citename{Alemany-Puig \bgroup et al.\egroup
  }2022]{Alemany2022a}
Llu\'is Alemany-Puig, {Juan Luis} Esteban, and Ramon {Ferrer-i-Cancho}.
\newblock 2022.
\newblock Minimum projective linearizations of trees in linear time.
\newblock {\em Information Processing Letters}, 174:106204.

\bibitem[\protect\citename{Alzetta \bgroup et al.\egroup }2017]{Alzetta2017a}
Chiara Alzetta, Felice Dell{'}Orletta, Simonetta Montemagni, and Giulia
  Venturi.
\newblock 2017.
\newblock Dangerous relations in dependency treebanks.
\newblock In {\em Proceedings of the 16th International Workshop on Treebanks
  and Linguistic Theories}, pages 201--210, Prague, Czech Republic.

\bibitem[\protect\citename{Alzetta \bgroup et al.\egroup }2018]{Alzetta2018a}
Chiara Alzetta, Felice Dell{'}Orletta, Simonetta Montemagni, and Giulia
  Venturi.
\newblock 2018.
\newblock {U}niversal {D}ependencies and quantitative typological trends. a
  case study on word order.
\newblock In {\em Proceedings of the Eleventh International Conference on
  Language Resources and Evaluation ({LREC} 2018)}, Miyazaki, Japan, May.
  European Language Resources Association (ELRA).

\bibitem[\protect\citename{Bennett \bgroup et al.\egroup }2019]{Bennett2019a}
Patrick Bennett, Sean English, and Maria {Talanda-Fisher}.
\newblock 2019.
\newblock Weighted {T}ur\'{a}n problems with applications.
\newblock {\em Discrete Mathematics}, 342:2165--2172, 8.

\bibitem[\protect\citename{Best and Rottmann}2017]{Best2017a}
{Karl-Heinz} Best and Otto Rottmann.
\newblock 2017.
\newblock {\em {Q}uantitative {L}inguistics, an {I}nvitation}.
\newblock RAM-Verlag, L\"{u}denscheid, Germany, 1 edition.

\bibitem[\protect\citename{Beyer and Hedetniemi}1980]{Beyer1980a}
Terry Beyer and {Sandra Mitchell} Hedetniemi.
\newblock 1980.
\newblock {Constant time generation of rooted trees}.
\newblock {\em SIAM Journal on Computing}, 9(4):706--712.

\bibitem[\protect\citename{Bommasani}2020]{Bommasani2020a}
Rishi Bommasani.
\newblock 2020.
\newblock Generalized optimal linear orders.
\newblock Master's thesis, Cornell University.

\bibitem[\protect\citename{Buchholz and Marsi}2006]{Buchholz2006a}
Sabine Buchholz and Erwin Marsi.
\newblock 2006.
\newblock {C}o{NLL}-{X} shared task on multilingual dependency parsing.
\newblock In {\em Proceedings of the Tenth Conference on Computational Natural
  Language Learning ({C}o{NLL}-X)}, pages 149--164, New York City, 06.
  Association for Computational Linguistics.

\bibitem[\protect\citename{Chung}1984]{Chung1984a}
{Fan R. K.} Chung.
\newblock 1984.
\newblock On optimal linear arrangements of trees.
\newblock {\em Computers \& Mathematics with Applications}, 10(1):43--60.

\bibitem[\protect\citename{Cordella \bgroup et al.\egroup }2004]{Cordella2004a}
{Luigi P.} Cordella, Pasquale Foggia, Carlo Sansone, and Mario Vento.
\newblock 2004.
\newblock A (sub)graph isomorphism algorithm for matching large graphs.
\newblock {\em IEEE Transactions on Pattern Analysis and Machine Intelligence},
  26(10):1367--1372.

\bibitem[\protect\citename{Courtin and Yan}2019]{Yan2019a}
Marine Courtin and Chunxiao Yan.
\newblock 2019.
\newblock What can we learn from natural and artificial dependency trees.
\newblock In {\em Proceedings of the First Workshop on Quantitative Syntax
  (Quasy, SyntaxFest 2019)}, pages 125--135, Paris, France, 08. Association for
  Computational Linguistics.

\bibitem[\protect\citename{Croft \bgroup et al.\egroup }2017]{Croft2017a}
William Croft, Dawn Nordquist, Katherine Looney, and Michael Regan.
\newblock 2017.
\newblock Linguistic typology meets universal dependencies.
\newblock In {\em TLT}.

\bibitem[\protect\citename{{de Marneffe} \bgroup et al.\egroup
  }2014]{StanfordDependencies}
{Marie-Catherine} {de Marneffe}, Timothy Dozat, Natalia Silveira, Katri
  Haverinen, Filip Ginter, Joakim Nivre, and {Christopher D.} Manning.
\newblock 2014.
\newblock Universal stanford dependencies: A cross-linguistic typology.
\newblock In {\em LREC}.

\bibitem[\protect\citename{DeVos and Nurse}2018]{Nurse2008a}
Matt DeVos and Kathryn Nurse.
\newblock 2018.
\newblock A maximum linear arrangement problem on directed graphs.
\newblock {\em arXiv}.

\bibitem[\protect\citename{Esteban and {Ferrer-i-Cancho}}2017]{Esteban2017a}
{Juan Luis} Esteban and Ramon {Ferrer-i-Cancho}.
\newblock 2017.
\newblock A correction on shiloach's algorithm for minimum linear arrangement
  of trees.
\newblock {\em SIAM Journal on Computing}, 46(3):1146--1151.

\bibitem[\protect\citename{Esteban \bgroup et al.\egroup }2016]{Esteban2016a}
{Juan Luis} Esteban, Ramon {Ferrer-i-Cancho}, and Carlos
  {G\'{o}mez-Rodr\'{i}guez}.
\newblock 2016.
\newblock The scaling of the minimum sum of edge lengths in uniformly random
  trees.
\newblock {\em Journal of Statistical Mechanics: Theory and Experiment},
  2016(6):063401, jun.

\bibitem[\protect\citename{{Ferrer-i-Cancho} and
  {G\'{o}mez-Rodr\'{i}guez}}2021]{Ferrer2019b}
Ramon {Ferrer-i-Cancho} and Carlos {G\'{o}mez-Rodr\'{i}guez}.
\newblock 2021.
\newblock Anti dependency distance minimization in short sequences. a graph
  theoretic approach.
\newblock {\em Journal of Quantitative Linguistics}, 28(1):50--76.

\bibitem[\protect\citename{{Ferrer-i-Cancho} \bgroup et al.\egroup
  }2018]{Ferrer2018a}
Ramon {Ferrer-i-Cancho}, Carlos {G\'omez-Rodr{\'i}guez}, and {Juan Luis}
  Esteban.
\newblock 2018.
\newblock Are crossing dependencies really scarce?
\newblock {\em Physica A: Statistical Mechanics and its Applications},
  493:311--329.

\bibitem[\protect\citename{{Ferrer-i-Cancho} \bgroup et al.\egroup
  }2021]{Ferrer2022a}
Ramon {Ferrer-i-Cancho}, Carlos {G\'{o}mez-Rodr\'{i}guez}, {Juan Luis} Esteban,
  and Llu\'{i}s {Alemany-Puig}.
\newblock 2021.
\newblock The optimality of syntactic dependency distances.
\newblock {\em Physical Review E}, page in press.

\bibitem[\protect\citename{{Ferrer-i-Cancho}}2003]{Ferrer2003a}
Ramon {Ferrer-i-Cancho}.
\newblock 2003.
\newblock {\em Language Universals: Principles and origins}.
\newblock {Ph.D.} thesis, Universitat Polit\'{e}cnica de Catalunya -
  BarcelonaTech.
\newblock (unpublished).

\bibitem[\protect\citename{{Ferrer-i-Cancho}}2004]{Ferrer2004a}
Ramon {Ferrer-i-Cancho}.
\newblock 2004.
\newblock Euclidean distance between syntactically linked words.
\newblock {\em Physical Review E}, 70(5):5.

\bibitem[\protect\citename{{Ferrer-i-Cancho}}2014]{Ferrer2014a}
Ramon {Ferrer-i-Cancho}.
\newblock 2014.
\newblock A stronger null hypothesis for crossing dependencies.
\newblock {\em {EPL} (Europhysics Letters)}, 108(5):58003, 12.

\bibitem[\protect\citename{{Ferrer-i-Cancho}}2019]{Ferrer2019a}
Ramon {Ferrer-i-Cancho}.
\newblock 2019.
\newblock The sum of edge lengths in random linear arrangements.
\newblock {\em Journal of Statistical Mechanics: Theory and Experiment},
  2019:053401, 05.

\bibitem[\protect\citename{Futrell \bgroup et al.\egroup }2015]{Futrell2015a}
Richard Futrell, Kyle Mahowald, and Edward Gibson.
\newblock 2015.
\newblock Large-scale evidence of dependency length minimization in 37
  languages.
\newblock {\em Proceedings of the National Academy of Sciences},
  112(33):10336--10341.

\bibitem[\protect\citename{Futrell \bgroup et al.\egroup }2020]{Futrell2020a}
Richard Futrell, {Roger Park} Levy, and Edward Gibson.
\newblock 2020.
\newblock Dependency locality as an explanatory principle for word order.
\newblock {\em Language}, 96(2):371--412.

\bibitem[\protect\citename{Garey and Johnson}1976]{Garey1976a}
{Michael R.} Garey and {David Stifler} Johnson.
\newblock 1976.
\newblock Some simplified {NP}-complete graph problems.
\newblock {\em Theoretical Computer Science}, pages 237--267.

\bibitem[\protect\citename{Gerdes \bgroup et al.\egroup
  }2018]{SurfaceUniversalDependencies2018}
Kim Gerdes, Bruno Guillaume, Sylvain Kahane, and Guy Perrier.
\newblock 2018.
\newblock {SUD} or surface-syntactic universal dependencies: An annotation
  scheme near-isomorphic to {UD}.
\newblock In {\em Proceedings of the Second Workshop on Universal Dependencies
  ({UDW} 2018)}, pages 66--74, Brussels, Belgium, 11. Association for
  Computational Linguistics.

\bibitem[\protect\citename{Gildea and Temperley}2007]{Gildea2007a}
Daniel Gildea and David Temperley.
\newblock 2007.
\newblock Optimizing grammars for minimum dependency length.
\newblock In {\em Proceedings of the 45th Annual Meeting of the Association of
  Computational Linguistics}, pages 184--191, Prague, Czech Republic, 06.
  Association for Computational Linguistics.

\bibitem[\protect\citename{Gildea and Temperley}2010]{Gildea2010a}
David Gildea and David Temperley.
\newblock 2010.
\newblock {D}o {G}rammars {M}inimize {D}ependency {L}ength?
\newblock {\em Cognitive Science}, 34(2):286--310.

\bibitem[\protect\citename{GMP}2021]{Web_GMP}
GMP.
\newblock 2021.
\newblock The {GNU} multiple precision arithmetic library.
\newblock \url{https://gmplib.org/}.
\newblock Accessed: 2021-09-16.

\bibitem[\protect\citename{Goldberg and Klipker}1976]{Goldberg1976a}
{Mark K.} Goldberg and {Israel A.} Klipker.
\newblock 1976.
\newblock A {M}inimal {P}lacement of a {T}ree on the {L}ine.
\newblock Technical report, Physico-Technical Institute of Low Temperatures.
  Academy of Sciences of Ukranian SSR, USSR.
\newblock in Russian.

\bibitem[\protect\citename{{G\'{o}mez-Rodr\'{i}guez} and
  {Ferrer-i-Cancho}}2017]{Gomez2016a}
Carlos {G\'{o}mez-Rodr\'{i}guez} and Ramon {Ferrer-i-Cancho}.
\newblock 2017.
\newblock Scarcity of crossing dependencies: {A} direct outcome of a specific
  constraint?
\newblock {\em Physical Review E}, 96:062304.

\bibitem[\protect\citename{{G\'{o}mez-Rodr\'{i}guez} \bgroup et al.\egroup
  }2011]{Gomez2011a}
Carlos {G\'{o}mez-Rodr\'{i}guez}, John Carroll, and David Weir.
\newblock 2011.
\newblock {D}ependency {P}arsing {S}chemata and {M}ildly {N}on-{P}rojective
  {D}ependency {P}arsing.
\newblock {\em Computational Linguistics}, 37(3):541--586.

\bibitem[\protect\citename{{G\'{o}mez-Rodr\'{i}guez} \bgroup et al.\egroup
  }2020]{Gomez2019a}
Carlos {G\'{o}mez-Rodr\'{i}guez}, {Morten H.} Christiansen, and Ramon
  {Ferrer-i-Cancho}.
\newblock 2020.
\newblock Memory limitations are hidden in grammar.
\newblock {\em Arxiv}, page under review.

\bibitem[\protect\citename{Gulordava and Merlo}2015]{Gulordava2015a}
Kristina Gulordava and Paola Merlo.
\newblock 2015.
\newblock Diachronic trends in word order freedom and dependency length in
  dependency-annotated corpora of {Latin} and ancient {Greek}.
\newblock In {\em Proceedings of the {Third International Conference on
  Dependency Linguistics (Depling 2015)}}, pages 121--130, Uppsala, Sweden, 08.
  Uppsala University.

\bibitem[\protect\citename{Gulordava and Merlo}2016]{Gulordava2016a}
Kristina Gulordava and Paola Merlo.
\newblock 2016.
\newblock Multi-lingual dependency parsing evaluation: a large-scale analysis
  of word order properties using artificial data.
\newblock {\em Transactions of the Association for Computational Linguistics},
  4:343--356.

\bibitem[\protect\citename{Haji\v{c} \bgroup et al.\egroup
  }2006]{PragueDependencies}
Jan Haji\v{c}, Jarmila Panevov\'{a}, Eva Haji\v{c}ov\'{a}, Jarmila
  Panevov\'{a}, Petr Sgall, Petr Pajas, Jan \v{S}t\v{e}p\'{a}nek, Ji\v{r}\'{\i}
  Havelka, and Marie Mikulov\'{a}.
\newblock 2006.
\newblock {Prague Dependency Treebank} {2.0}.
\newblock CDROM CAT: LDC2006T01, ISBN 1-58563-370-4. Linguistic Data
  Consortium.

\bibitem[\protect\citename{Harary and Schwenk}1973]{Harary1973a}
Frank Harary and {Allen J.} Schwenk.
\newblock 1973.
\newblock The number of caterpillars.
\newblock {\em Discrete Mathematics}, 6:359--365.

\bibitem[\protect\citename{Harary}1969]{Harary1969a}
Frank Harary.
\newblock 1969.
\newblock {\em Graph Theory}.
\newblock Addison-Wesley, Reading, MA.

\bibitem[\protect\citename{Hassin and Rubinstein}2000]{Hassin2000a}
Refael Hassin and Shlomi Rubinstein.
\newblock 2000.
\newblock Approximation algorithms for maximum linear arrangement.
\newblock In {\em Scandinavian Workshop on Algorithm Theory - Algorithm Theory
  - SWAT 2000}, volume 1851, pages 231--236.

\bibitem[\protect\citename{Heinecke}2019]{Heinecke2019a}
Johannes Heinecke.
\newblock 2019.
\newblock {C}onllu{E}ditor: a fully graphical editor for universal dependencies
  treebank files.
\newblock In {\em Proceedings of the Third Workshop on Universal Dependencies
  (UDW, SyntaxFest 2019)}, pages 87--93, Paris, France, 08. Association for
  Computational Linguistics.

\bibitem[\protect\citename{Hochberg and Stallmann}2003]{Hochberg2003a}
{Robert A.} Hochberg and {Matthias F.} Stallmann.
\newblock 2003.
\newblock Optimal one-page tree embeddings in linear time.
\newblock {\em Information Processing Letters}, 87(2):59--66.

\bibitem[\protect\citename{Hudson}1995]{Hudson1995a}
Richard Hudson.
\newblock 1995.
\newblock Measuring syntactic difficulty.
\newblock {\em Unpublished paper}.

\bibitem[\protect\citename{Iordanskii}1987]{Iordanskii1987a}
{Mikhail Anatolievich} Iordanskii.
\newblock 1987.
\newblock Minimal numberings of the vertices of trees --- approximate approach.
\newblock In Lothar Budach, Rais~Gati{\v{c}} Bukharajev, and
  Oleg~Borisovi{\v{c}} Lupanov, editors, {\em Fundamentals of Computation
  Theory}, pages 214--217, Berlin, Heidelberg. Springer Berlin Heidelberg.

\bibitem[\protect\citename{Jiang and Liu}2018]{Jiang2018a}
Jingyang Jiang and Haitao Liu, editors.
\newblock 2018.
\newblock {\em {Q}uantitative {A}nalysis of {D}ependency {S}tructures}.
\newblock De Gruyter Mouton, Berlin, 1 edition.

\bibitem[\protect\citename{Jing and Liu}2015]{Jing2015a}
Yingqi Jing and Haitao Liu.
\newblock 2015.
\newblock Mean hierarchical distance. {Augmenting} mean dependency distance.
\newblock In {\em Proceedings of the Third International Conference on
  Dependency Linguistics}, pages 161--170.

\bibitem[\protect\citename{J\"{u}ttner and Madarasi}2018]{Juttner2018a}
Alp\'{a}r J\"{u}ttner and P\'{e}ter Madarasi.
\newblock 2018.
\newblock {VF2++} -- {A}n improved subgraph isomorphism algorithm.
\newblock {\em Discrete Applied Mathematics}, 242:69--81.
\newblock Computational Advances in Combinatorial Optimization.

\bibitem[\protect\citename{Kahane \bgroup et al.\egroup }2017]{Kahane2017a}
Sylvain Kahane, Chunxiao Yan, and {Marie-Am\'{e}lie} Botalla.
\newblock 2017.
\newblock What are the limitations on the flux of syntactic dependencies?
  evidence from ud treebanks.
\newblock In {\em Proceedings of the Fourth International Conference on
  Dependency Linguistics}, pages 73--82, 9.

\bibitem[\protect\citename{{K\"{o}hler} and Altmann}2012]{Kholer2012a}
Reinhard {K\"{o}hler} and Gabriel Altmann.
\newblock 2012.
\newblock {\em Quantitative Syntax Analysis}.
\newblock Quantitative linguistics. De Gruyter Mouton.

\bibitem[\protect\citename{Komori \bgroup et al.\egroup }2019]{Komori2019a}
Saeko Komori, Masatoshi Sugiura, and Wenping Li.
\newblock 2019.
\newblock Examining {MDD} and {MHD} as syntactic complexity measures with
  intermediate {J}apanese learner corpus data.
\newblock In {\em Proceedings of the Fifth International Conference on
  Dependency Linguistics (Depling, SyntaxFest 2019)}, pages 130--135, Paris,
  France, 08. Association for Computational Linguistics.

\bibitem[\protect\citename{Kramer}2021]{Kramer2021a}
Alex Kramer.
\newblock 2021.
\newblock Dependency lengths in speech and writing: A cross-linguistic
  comparison via {YouDePP}, a pipeline for scraping and parsing {YouTube}
  captions.
\newblock In {\em Proceedings of the Society for Computation in Linguistics},
  volume~4, pages 359--365.

\bibitem[\protect\citename{Kuhlmann and Nivre}2006]{Kuhlmann2006a}
Marco Kuhlmann and Joakim Nivre.
\newblock 2006.
\newblock Mildly non-projective dependency structures.
\newblock In {\em Proceedings of the COLING/ACL 2006 Main Conference Poster
  Sessions}, COLING-ACL '06, pages 507--514, 07.

\bibitem[\protect\citename{Liu \bgroup et al.\egroup }2017]{Liu2017a}
Haitao Liu, Chunshan Xu, and Junying Liang.
\newblock 2017.
\newblock Dependency distance: A new perspective on syntactic patterns in
  natural languages.
\newblock {\em Physics of Life Reviews}, 21:171--193.

\bibitem[\protect\citename{Liu}2007]{Liu2007a}
Haitao Liu.
\newblock 2007.
\newblock Probability distribution of dependency distance.
\newblock {\em Glottometrics}, 15:1--12, 06.

\bibitem[\protect\citename{Liu}2008]{Liu2008a}
Haitao Liu.
\newblock 2008.
\newblock Dependency distance as a metric of language comprehension difficulty.
\newblock {\em Journal of Cognitive Science}, 9(2):159--191.

\bibitem[\protect\citename{Liu}2010]{Liu2010a}
Haitao Liu.
\newblock 2010.
\newblock Dependency direction as a means of word-order typology: a method
  based on dependency treebanks.
\newblock {\em Lingua}, 120(6):1567--1578.

\bibitem[\protect\citename{Marohnić}2018]{GiacXcas_Manual}
Luka Marohnić.
\newblock 2018.
\newblock Graph theory package for giac/xcas - reference manual.
\newblock \url{
  https://usermanual.wiki/Document/graphtheoryusermanual.346702481/view}.
\newblock Accessed: 2020-01-13.

\bibitem[\protect\citename{McDonald \bgroup et al.\egroup }2005]{McDonald2005a}
Ryan McDonald, Fernando Pereira, Kiril Ribarov, and Jan Haji\v{c}.
\newblock 2005.
\newblock Non-projective dependency parsing using spanning tree algorithms.
\newblock In {\em Proceedings of the conference on Human Language Technology
  and Empirical Methods in Natural Language Processing}, pages 523--530.

\bibitem[\protect\citename{Mel'{\v{c}}uk}1988]{Melcuk1988a}
Igor Mel'{\v{c}}uk.
\newblock 1988.
\newblock {\em {D}ependency {S}yntax: {T}heory and {P}ractice}.
\newblock State University of New York Press, Albany, NY, USA.

\bibitem[\protect\citename{Nijenhuis and Wilf}1978]{Nijenhuis1978a}
Albert Nijenhuis and {Herbert S.} Wilf.
\newblock 1978.
\newblock {\em Combinatorial Algorithms: For Computers and Hard Calculators}.
\newblock Academic Press, Inc., Orlando, FL, USA, 2nd edition.

\bibitem[\protect\citename{Nivre}2006]{Nivre2006a}
Joakim Nivre.
\newblock 2006.
\newblock Constraints on non-projective dependency parsing.
\newblock In {\em EACL 2006 - 11th Conference of the European Chapter of the
  Association for Computational Linguistics, Proceedings of the Conference},
  pages 73--80.

\bibitem[\protect\citename{Osborne and Gerdes}2019]{Osborne2019a}
T.~Osborne and K.~Gerdes.
\newblock 2019.
\newblock The status of function words in dependency grammar: {A} critique of
  {Universal Dependencies (UD)}.
\newblock {\em Glossa: A Journal of General Linguistics}, 4(1):17.

\bibitem[\protect\citename{Park and Levy}2009]{Park2009a}
{Y. Albert} Park and {Roger Park} Levy.
\newblock 2009.
\newblock Minimal-length linearizations for mildly context-sensitive dependency
  trees.
\newblock In {\em Proceedings of the {10th Annual Meeting of the North American
  Chapter of the Association for Computational Linguistics: Human Language
  Technologies (NAACL-HLT)} conference}, pages 335--343, Stroudsburg, PA, USA.
  Association for Computational Linguistics.

\bibitem[\protect\citename{Passarotti}2016]{Passarotti2016a}
{Marco Carlo} Passarotti.
\newblock 2016.
\newblock How far is stanford from prague (and vice versa)? comparing two
  dependency-based annotation schemes by network analysis.
\newblock {\em L'ANALISI LINGUISTICA E LETTERARIA}, pages 21--46.

\bibitem[\protect\citename{Pighin}2012]{Manual_TikzDep}
Daniele Pighin.
\newblock 2012.
\newblock The {Ti$k$Z}-{\tt dependency} package.
\newblock
  \url{https://osl.ugr.es/CTAN/graphics/pgf/contrib/tikz-dependency/tikz-dependency-doc.pdf}.
\newblock Accessed: 2021-06-17.

\bibitem[\protect\citename{Pr\"ufer}1918]{Pruefer1918a}
Heinz Pr\"ufer.
\newblock 1918.
\newblock {Neuer Beweis eines Satzes \"uber Permutationen}.
\newblock {\em Arch. Math. Phys}, 27:742--744.

\bibitem[\protect\citename{R\'{e}nyi and Szekeres}1967]{Renyi1965a}
Alfr\'{e}d R\'{e}nyi and George Szekeres.
\newblock 1967.
\newblock On the height of trees.
\newblock {\em Journal of the Australian Mathematical Society}, 7(4):497–507.

\bibitem[\protect\citename{{San Diego} and Gella}2014]{SanDiego2014a}
{Immanuel T.} {San Diego} and {Frederick S.} Gella.
\newblock 2014.
\newblock The $b$-chromatic number of bistar graph.
\newblock {\em Applied Mathematical Sciences}, 8(116):5795--5800.

\bibitem[\protect\citename{Satta \bgroup et al.\egroup }2013]{Pitler2013a}
Giorgio Satta, Emily Pitler, Sampath Kannan, and Mitchell Marcus.
\newblock 2013.
\newblock Finding optimal 1-{E}ndpoint-{C}rossing trees.
\newblock In {\em Transactions of the Association for Computational
  Linguistics}, pages 13--24.

\bibitem[\protect\citename{Shiloach}1979]{Shiloach1979a}
Yossi Shiloach.
\newblock 1979.
\newblock A minimum linear arrangement algorithm for undirected trees.
\newblock {\em SIAM Journal on Computing}, 8(1):15--32.

\bibitem[\protect\citename{Sleator and Temperley}1993]{Sleator1993a}
Daniel Sleator and Davy Temperley.
\newblock 1993.
\newblock Parsing {English} with a link grammar.
\newblock In {\em Proceedings of the Third International Workshop on Parsing
  Technologies (IWPT’93)}, pages 277--292. ACL/SIGPARSE.

\bibitem[\protect\citename{SWIG}2020]{Web_SWIG}
SWIG.
\newblock 2020.
\newblock Swig 4.0.2.
\newblock \url{http://www.swig.org/}.
\newblock Accessed: 2021-06-09.

\bibitem[\protect\citename{Temperley and Gildea}2018]{Temperley2018a}
David Temperley and Daniel Gildea.
\newblock 2018.
\newblock Minimizing syntactic dependency lengths: Typological/cognitive
  universal?
\newblock {\em Annual Review of Linguistics}, 4(1):67--80.

\bibitem[\protect\citename{Verbitsky}2008]{Verbitsky2008a}
Oleg Verbitsky.
\newblock 2008.
\newblock On the obfuscation complexity of planar graphs.
\newblock {\em Theoretical Computer Science}, 396(1):294--300.

\bibitem[\protect\citename{Wilf}1981]{Wilf1981a}
{Herbert S.} Wilf.
\newblock 1981.
\newblock The uniform selection of free trees.
\newblock {\em Journal of Algorithms}, 2:204--207.

\bibitem[\protect\citename{Wright \bgroup et al.\egroup }1986]{Wright1986a}
Robert~Alan Wright, Bruce Richmond, Andrew Odlyzko, and {Brendan D.} McKay.
\newblock 1986.
\newblock Constant time generation of free trees.
\newblock {\em SIAM Journal on Computing}, 15:540--548, 05.

\bibitem[\protect\citename{Yadav \bgroup et al.\egroup }2019]{Yadav2019a}
Himanshu Yadav, Samar Husain, and Richard Futrell.
\newblock 2019.
\newblock Are formal restrictions on crossing dependencies epiphenominal?
\newblock In {\em Proceedings of the 18th International Workshop on Treebanks
  and Linguistic Theories (TLT, SyntaxFest 2019)}, pages 2--12, Paris, France,
  08. Association for Computational Linguistics.

\bibitem[\protect\citename{Yadav \bgroup et al.\egroup }2021]{Yadav2021a}
Himanshu Yadav, Samar Husain, and Richard Futrell.
\newblock 2021.
\newblock Do dependency lengths explain constraints on crossing dependencies?
\newblock {\em Linguistics Vanguard}, 7(s3):20190070.

\bibitem[\protect\citename{Yu \bgroup et al.\egroup }2019]{Yu2019a}
Xiang Yu, Agnieszka Falenska, and Jonas Kuhn.
\newblock 2019.
\newblock Dependency length minimization vs. word order constraints: An
  empirical study on 55 treebanks.
\newblock In {\em Proceedings of the First Workshop on Quantitative Syntax
  (Quasy, SyntaxFest 2019)}, pages 89--97, Paris, France, 08. Association for
  Computational Linguistics.

\bibitem[\protect\citename{Zeman \bgroup et al.\egroup
  }2020]{UniversalDependencies26}
Daniel Zeman, Joakim Nivre, Mitchell Abrams, Elia Ackermann, No{\"e}mi Aepli,
  {\v Z}eljko Agi{\'c}, Lars Ahrenberg, Chika~Kennedy Ajede, Gabriel{\.e}
  Aleksandravi{\v c}i{\=u}t{\.e}, Lene Antonsen, Katya Aplonova, Angelina
  Aquino, Maria~Jesus Aranzabe, Gashaw Arutie, Masayuki Asahara, Luma Ateyah,
  Furkan Atmaca, Mohammed Attia, Aitziber Atutxa, Liesbeth Augustinus, Elena
  Badmaeva, Miguel Ballesteros, Esha Banerjee, Sebastian Bank, Verginica
  Barbu~Mititelu, Victoria Basmov, Colin Batchelor, John Bauer, Kepa
  Bengoetxea, Yevgeni Berzak, Irshad~Ahmad Bhat, Riyaz~Ahmad Bhat, Erica
  Biagetti, Eckhard Bick, Agn{\.e} Bielinskien{\.e}, Rogier Blokland, Victoria
  Bobicev, Lo{\"{\i}}c Boizou, Emanuel Borges~V{\"o}lker, Carl B{\"o}rstell,
  Cristina Bosco, Gosse Bouma, Sam Bowman, Adriane Boyd, Kristina Brokait{\.e},
  Aljoscha Burchardt, Marie Candito, Bernard Caron, Gauthier Caron, Tatiana
  Cavalcanti, G{\"u}l{\c s}en Cebiro{\u g}lu~Eryi{\u g}it, Flavio~Massimiliano
  Cecchini, Giuseppe G.~A. Celano, Slavom{\'{\i}}r {\v C}{\'e}pl{\"o}, Savas
  Cetin, Fabricio Chalub, Ethan Chi, Jinho Choi, Yongseok Cho, Jayeol Chun,
  Alessandra~T. Cignarella, Silvie Cinkov{\'a}, Aur{\'e}lie Collomb, {\c C}a{\u
  g}r{\i} {\c C}{\"o}ltekin, Miriam Connor, Marine Courtin, Elizabeth Davidson,
  Marie-Catherine de~Marneffe, Valeria de~Paiva, Elvis de~Souza, Arantza
  Diaz~de Ilarraza, Carly Dickerson, Bamba Dione, Peter Dirix, Kaja Dobrovoljc,
  Timothy Dozat, Kira Droganova, Puneet Dwivedi, Hanne Eckhoff, Marhaba Eli,
  Ali Elkahky, Binyam Ephrem, Olga Erina, Toma{\v z} Erjavec, Aline Etienne,
  Wograine Evelyn, Rich{\'a}rd Farkas, Hector Fernandez~Alcalde, Jennifer
  Foster, Cl{\'a}udia Freitas, Kazunori Fujita, Katar{\'{\i}}na Gajdo{\v
  s}ov{\'a}, Daniel Galbraith, Marcos Garcia, Moa G{\"a}rdenfors, Sebastian
  Garza, Kim Gerdes, Filip Ginter, Iakes Goenaga, Koldo Gojenola, Memduh
  G{\"o}k{\i}rmak, Yoav Goldberg, Xavier G{\'o}mez~Guinovart, Berta
  Gonz{\'a}lez~Saavedra, Bernadeta Grici{\=u}t{\.e}, Matias Grioni, Lo{\"{\i}}c
  Grobol, Normunds Gr{\=u}z{\={\i}}tis, Bruno Guillaume, C{\'e}line
  Guillot-Barbance, Tunga G{\"u}ng{\"o}r, Nizar Habash, Jan Haji{\v c}, Jan
  Haji{\v c}~jr., Mika H{\"a}m{\"a}l{\"a}inen, Linh H{\`a}~M{\~y}, Na-Rae Han,
  Kim Harris, Dag Haug, Johannes Heinecke, Oliver Hellwig, Felix Hennig,
  Barbora Hladk{\'a}, Jaroslava Hlav{\'a}{\v c}ov{\'a}, Florinel Hociung,
  Petter Hohle, Jena Hwang, Takumi Ikeda, Radu Ion, Elena Irimia, {\d
  O}l{\'a}j{\'{\i}}d{\'e} Ishola, Tom{\'a}{\v s} Jel{\'{\i}}nek, Anders
  Johannsen, Hildur J{\'o}nsd{\'o}ttir, Fredrik J{\o}rgensen, Markus Juutinen,
  H{\"u}ner Ka{\c s}{\i}kara, Andre Kaasen, Nadezhda Kabaeva, Sylvain Kahane,
  Hiroshi Kanayama, Jenna Kanerva, Boris Katz, Tolga Kayadelen, Jessica Kenney,
  V{\'a}clava Kettnerov{\'a}, Jesse Kirchner, Elena Klementieva, Arne K{\"o}hn,
  Abdullatif K{\"o}ksal, Kamil Kopacewicz, Timo Korkiakangas, Natalia Kotsyba,
  Jolanta Kovalevskait{\.e}, Simon Krek, Sookyoung Kwak, Veronika Laippala,
  Lorenzo Lambertino, Lucia Lam, Tatiana Lando, Septina~Dian Larasati, Alexei
  Lavrentiev, John Lee, Phuong L{\^e}~H{\`{\^o}}ng, Alessandro Lenci, Saran
  Lertpradit, Herman Leung, Maria Levina, Cheuk~Ying Li, Josie Li, Keying Li,
  {KyungTae} Lim, Yuan Li, Nikola Ljube{\v s}i{\'c}, Olga Loginova, Olga
  Lyashevskaya, Teresa Lynn, Vivien Macketanz, Aibek Makazhanov, Michael Mandl,
  Christopher Manning, Ruli Manurung, C{\u a}t{\u a}lina M{\u a}r{\u a}nduc,
  David Mare{\v c}ek, Katrin Marheinecke, H{\'e}ctor Mart{\'{\i}}nez~Alonso,
  Andr{\'e} Martins, Jan Ma{\v s}ek, Hiroshi Matsuda, Yuji Matsumoto, Ryan
  {McDonald}, Sarah {McGuinness}, Gustavo Mendon{\c c}a, Niko Miekka, Margarita
  Misirpashayeva, Anna Missil{\"a}, C{\u a}t{\u a}lin Mititelu, Maria Mitrofan,
  Yusuke Miyao, Simonetta Montemagni, Amir More, Laura Moreno~Romero,
  Keiko~Sophie Mori, Tomohiko Morioka, Shinsuke Mori, Shigeki Moro, Bjartur
  Mortensen, Bohdan Moskalevskyi, Kadri Muischnek, Robert Munro, Yugo Murawaki,
  Kaili M{\"u}{\"u}risep, Pinkey Nainwani, Juan~Ignacio Navarro~Hor{\~n}iacek,
  Anna Nedoluzhko, Gunta Ne{\v s}pore-B{\=e}rzkalne, Luong Nguy{\~{\^e}}n~Th{\d
  i}, Huy{\`{\^e}}n Nguy{\~{\^e}}n Th{\d i}~Minh, Yoshihiro Nikaido, Vitaly
  Nikolaev, Rattima Nitisaroj, Hanna Nurmi, Stina Ojala, Atul~Kr. Ojha,
  Ad{\'e}day{\d o} Ol{\'u}{\`o}kun, Mai Omura, Emeka Onwuegbuzia, Petya
  Osenova, Robert {\"O}stling, Lilja {\O}vrelid, {\c S}aziye~Bet{\"u}l
  {\"O}zate{\c s}, Arzucan {\"O}zg{\"u}r, Balk{\i}z {\"O}zt{\"u}rk~Ba{\c
  s}aran, Niko Partanen, Elena Pascual, Marco Passarotti, Agnieszka Patejuk,
  Guilherme Paulino-Passos, Angelika Peljak-{\L}api{\'n}ska, Siyao Peng,
  Cenel-Augusto Perez, Guy Perrier, Daria Petrova, Slav Petrov, Jason Phelan,
  Jussi Piitulainen, Tommi~A Pirinen, Emily Pitler, Barbara Plank, Thierry
  Poibeau, Larisa Ponomareva, Martin Popel, Lauma Pretkalni{\c n}a, Sophie
  Pr{\'e}vost, Prokopis Prokopidis, Adam Przepi{\'o}rkowski, Tiina Puolakainen,
  Sampo Pyysalo, Peng Qi, Andriela R{\"a}{\"a}bis, Alexandre Rademaker,
  Loganathan Ramasamy, Taraka Rama, Carlos Ramisch, Vinit Ravishankar, Livy
  Real, Petru Rebeja, Siva Reddy, Georg Rehm, Ivan Riabov, Michael Rie{\ss}ler,
  Erika Rimkut{\.e}, Larissa Rinaldi, Laura Rituma, Luisa Rocha, Mykhailo
  Romanenko, Rudolf Rosa, Valentin Roșca, Davide Rovati, Olga Rudina, Jack
  Rueter, Shoval Sadde, Beno{\^{\i}}t Sagot, Shadi Saleh, Alessio Salomoni,
  Tanja Samard{\v z}i{\'c}, Stephanie Samson, Manuela Sanguinetti, Dage
  S{\"a}rg, Baiba Saul{\={\i}}te, Yanin Sawanakunanon, Salvatore Scarlata,
  Nathan Schneider, Sebastian Schuster, Djam{\'e} Seddah, Wolfgang Seeker,
  Mojgan Seraji, Mo~Shen, Atsuko Shimada, Hiroyuki Shirasu, Muh Shohibussirri,
  Dmitry Sichinava, Aline Silveira, Natalia Silveira, Maria Simi, Radu
  Simionescu, Katalin Simk{\'o}, M{\'a}ria {\v S}imkov{\'a}, Kiril Simov, Maria
  Skachedubova, Aaron Smith, Isabela Soares-Bastos, Carolyn Spadine, Antonio
  Stella, Milan Straka, Jana Strnadov{\'a}, Alane Suhr, Umut Sulubacak, Shingo
  Suzuki, Zsolt Sz{\'a}nt{\'o}, Dima Taji, Yuta Takahashi, Fabio Tamburini,
  Takaaki Tanaka, Samson Tella, Isabelle Tellier, Guillaume Thomas, Liisi
  Torga, Marsida Toska, Trond Trosterud, Anna Trukhina, Reut Tsarfaty, Utku
  T{\"u}rk, Francis Tyers, Sumire Uematsu, Roman Untilov, Zde{\v n}ka Ure{\v
  s}ov{\'a}, Larraitz Uria, Hans Uszkoreit, Andrius Utka, Sowmya Vajjala,
  Daniel van Niekerk, Gertjan van Noord, Viktor Varga, Eric Villemonte de~la
  Clergerie, Veronika Vincze, Aya Wakasa, Lars Wallin, Abigail Walsh, Jing~Xian
  Wang, Jonathan~North Washington, Maximilan Wendt, Paul Widmer, Seyi Williams,
  Mats Wir{\'e}n, Christian Wittern, Tsegay Woldemariam, Tak-sum Wong, Alina
  Wr{\'o}blewska, Mary Yako, Kayo Yamashita, Naoki Yamazaki, Chunxiao Yan,
  Koichi Yasuoka, Marat~M. Yavrumyan, Zhuoran Yu, Zden{\v e}k {\v
  Z}abokrtsk{\'y}, Amir Zeldes, Hanzhi Zhu, and Anna Zhuravleva.
\newblock 2020.
\newblock Universal dependencies 2.6.
\newblock {LINDAT}/{CLARIAH}-{CZ} digital library at the Institute of Formal
  and Applied Linguistics ({{\'U}FAL}), Faculty of Mathematics and Physics,
  Charles University.

\bibitem[\protect\citename{Zipf}1949]{Zipf1949a}
{George Kingsley} Zipf.
\newblock 1949.
\newblock {\em Human Behavior and the Principle of Least Effort: An
  Introduction to Human Ecology}.
\newblock Addison-Wesley Press, Oxford, England.

\end{thebibliography}

\end{document}